%% file: root.tex
\documentclass[final,5p,times,twocolumn]{elsarticle}

\usepackage{amssymb}

\usepackage{epsfig}
\usepackage{graphicx}
\usepackage{amsmath}

\usepackage{algorithm}
\usepackage{algorithmic}
\usepackage[labelformat=simple]{subcaption}

\usepackage{booktabs}
\usepackage{breqn}
\usepackage{tabularx}

\usepackage{cite}
\usepackage{amsfonts}
\usepackage{textcomp}
\usepackage{balance}
\usepackage{tikz}
\usepackage{pgfplots}
\pgfplotsset{compat=1.17}
\usepackage{soul}

\usepackage{hyperref}
\usepackage{wrapfig}

\newlength\myindent
\newcommand{\etal}{{\textit{et al}}.\@}
\DeclareMathOperator*{\argmax}{\mathrm{arg\,max}}

\journal{Arxiv}

\begin{document}
\begin{frontmatter}

\title{Neural Mixture Models with Expectation-Maximization for End-to-end Deep Clustering}

\author[1,2]{Dumindu Tissera}
\corref{cor1}
\cortext[cor1]{Corresponding author: 
  Tel.: +94 711512777}
\ead{dumindutissera@gmail.com}

\author[2]{Kasun Vithanage}
\author[1,2]{Rukshan Wijesinghe}
\author[2]{Alex Xavier}
\author[2]{Sanath Jayasena}
\author[2]{Subha Fernando}

\author[1,2]{Ranga Rodrigo}
\ead{ranga@uom.lk}
\ead[url]{http://ranga.staff.uom.lk}

\address[1]{Department of Electronic \& Telecommunication Engineering, Univerisity of Moratuwa, Sri Lanka}
\address[2]{CodeGen QBITS Lab, University of Moratuwa, Sri Lanka}

\begin{abstract}
Any clustering algorithm must synchronously learn to model the clusters and allocate data to those clusters in the absence of labels. Mixture model-based methods model clusters with pre-defined statistical distributions and allocate data to those clusters based on the cluster likelihoods. They iteratively refine those distribution parameters and member assignments following the Expectation-Maximization (EM) algorithm. However, the cluster representability of such hand-designed distributions that employ a limited amount of parameters is not adequate for most real-world clustering tasks. In this paper, we realize mixture model-based clustering with a neural network where the final layer neurons, with the aid of an additional transformation, approximate cluster distribution outputs. The network parameters pose as the parameters of those distributions. The result is an elegant, much-generalized representation of clusters than a restricted mixture of hand-designed distributions. We train the network end-to-end via batch-wise EM iterations where the forward pass acts as the E-step and the backward pass acts as the M-step.  In image clustering, the mixture-based EM objective can be used as the clustering objective along with existing representation learning methods. In particular, we show that when mixture-EM optimization is fused with consistency optimization, it improves the sole consistency optimization performance in clustering. Our trained networks outperform single-stage deep clustering methods that still depend on k-means, with unsupervised classification accuracy of \textbf{63.8\%} in STL10, \textbf{58\%} in CIFAR10, \textbf{25.9\%} in CIFAR100, and \textbf{98.9\%} in MNIST.
\end{abstract}

\begin{keyword}
Deep Clustering \sep Mixture Models \sep Expectation-Maximization 
\end{keyword}

\end{frontmatter}

\section{Introduction}
\label{se:intro}
\begin{figure*}[t]
	\begin{center}
	\includegraphics[width=0.9\linewidth]{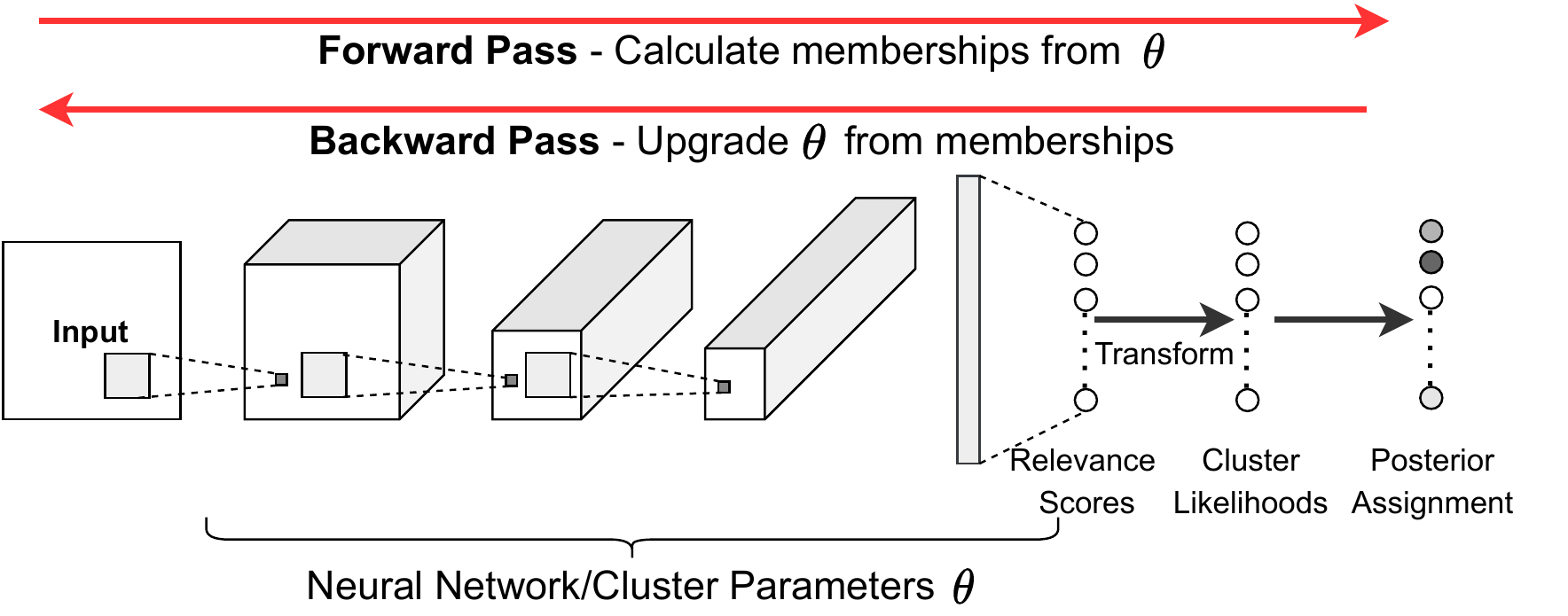}
	\end{center}
	\vspace{-0.2in}
	\caption{The expected outcome of Mixture-EM optimization on a neural network (NN) for clustering: The NN parameters $\theta$ represent the cluster parameters. In the forward pass, the NN outputs cluster relevance scores which are transformed to cluster distribution outputs/likelihoods. Using these likelihoods, the posterior probabilities of membership are derived for the existing parameters $\theta$. In the backward pass, the parameters $\theta$ get updated from the computed memberships.}
	\label{fig:intro}
\end{figure*}

Clustering identifies similarities among data points and groups similar data points together. Such automatic data grouping is significant when manually annotating classes is overwhelming, or human understanding is insufficient to annotate (e.g., non-image data). Usually, with the absence of labels, clustering algorithms have to learn both the cluster representations (modeled by parameters) and the member assignment. Most clustering algorithms do this by iteratively alternating between two steps: 
    1) assign samples (soft or hard) to clusters according to the degree of match to the cluster representations and,
    2) based on the cluster assignments, update current cluster representations.
This iterative nature shares the intuition of the Expectation-Maximization (EM) \citep{em} algorithm. 

Mixture models discover the existence of subpopulations within a given population. Such sub-populations can be used cumulatively to describe the properties of the whole population. In the clustering context, this can be thought of as discovering underlying homogeneous groups within a dataset and interpreting the dataset's properties from the discovered groups. Mixture models use hand-designed distributions (e.g., Gaussian or Bernoulli) with pre-defined parameters (e.g., a probability or mean and covariance) to represent clusters. A mixture of such distributions is fitted to the observed data points to maximize the total likelihood. This likelihood optimization is conducted via Expectation-Maximization, where the distribution parameters and posterior member assignments are iteratively refined, keeping one constant at a time. However, these statistical distributions contain a limited number of parameters to learn. Thus, traditional mixture models show poor cluster representations when directly used to cluster data with high dimensionality or high complexity. Since the strength of a clustering algorithm relies on its ability to represent clusters, we look into modeling advanced clusters with more freedom in the complexity of the discovered cluster distributions. Such distributions are not pre-defined and are adaptive to the dataset.

Neural Networks are universal approximators that can approximate any measurable function when provided with an adequate amount of learning, a sufficient number of hidden units, and a deterministic relationship between network inputs and outputs \citep{hornik1989multilayer}. Therefore, a NN is an ideal candidate for such advanced modeling of cluster distributions. If we employ a NN for such cluster modeling, we can let each final layer neuron approximate the relevance of a given observation (NN input) to a particular cluster. We can further transform such relevance to the respective cluster
distribution output/likelihood of the observation for that particular cluster. Once we have such cluster likelihoods for a batch of observations, we can calculate the posterior probabilities cluster assignment and formulate the EM objective as conventional mixture modeling. The EM objective can then be backpropagated to update the NN parameters. In this way, the NN parameters learn to represent shared and cluster-specific cluster parameters, leading to advanced cluster distributions. To our knowledge, an effort to employ a deep NN to directly model cluster distributions via their output nodes in mixture modeling has not been recorded yet.

This paper adopts EM-based mixture modeling as the clustering objective to train a NN end-to-end, where the NN itself models the cluster distributions. We formulate an EM-based learning algorithm in batch-wise iterations to learn the cluster representations and the cluster assignment concurrently. The contributions of this paper are three-fold. Firstly, we approximate each cluster distribution for a given batch by the parametric function of the network from the input to the corresponding final layer neuron, followed by an additional transformation. We impose such additional transformation to regulate the transformed final layer neuron to behave as a probability density function of the input. This prevents a single cluster distribution from overpowering other distributions to capture all datapoints, leading to the trivial solution of assigning all datapoints to that dominant cluster. The proposed transformation normalizes the output of each final layer neuron over the batch to have a zero batch-mean and further constraints to the most linear region of the $\mathrm{sigmoid}$ activation. The $\mathrm{sigmoid}$ activation maps its zero batch-mean linear region input to a continuous non-negative value, the distribution output. Constraining the final-layer neurons to have a zero batch-mean and within the most linear portion of the $\mathrm{sigmoid}$ enables all cluster distributions to share a common integral over the sample space, as shown later. Thus, the approximated distributions act as probability density functions of the observation, preventing trivial solutions.

Secondly, we propose a batch-wise EM optimization to train the network end-to-end. We calculate the posterior probabilities of member assignments using the approximated cluster likelihoods. As shown later, these posteriors are approximated by the $\mathrm{softmax}$ of the normalized final-layer neurons. We formulate the EM loss to backpropagate using the calculated posteriors and cluster likelihoods. The optimization process of the network performs EM iterations batch-wise in an online fashion. For each iteration, we feed a batch of observations to the network. The forward pass through the network corresponds to the E-step, where we calculate the cluster likelihoods and posterior probabilities from the NN output for the given batch of observations and derive the EM loss. The backward pass through the network corresponds to the M-step, where we perform a gradient step in optimizing the EM loss. Fig.~\ref{fig:intro} illustrates the overview of the proposed Mixture-EM optimization using a NN.
 
Thirdly, we integrate the EM optimization to consistency optimization between original and augmented datapoints for image clustering. It is essential to let the neural network learn general semantically important features and prevent the network from overfitting to only the lower-level information in data in clustering images. While this can be done by either learning a self-supervised pretext task \citep{dec} or learning to output a consistent model response to the original and its transform versions \citep{laine2016temporal, i2c, diallo2021deep}, we choose the latter and integrate it to our mixture-EM optimization. Our consistency optimization includes minimization of the Kullback–Leibler (KL) divergence \citep{kullback1951information} between the model responses to original images and their transformed versions. We embed this optimization into the EM process, resulting in a two-fold optimization. We show that this two-fold optimization shows accelerated and better convergence than the consistency optimization alone. 

Our framework does not use any conventional clustering techniques such as k-means. The proposed optimization performs EM iterations batch-wise in an online fashion, thus, eliminates the need to iterate over the entire sample space for a single update. Using a NN allows learning complex cluster representations rather than limiting them to hand-designed parameters. The transformation of the final layer neurons to regulate cluster distributions as probability density functions prevents the model from collapsing to trivial solutions. Thus, this simple transformation eliminates additional effort to force the model to divide the samples among the clusters evenly. The training process consists of only the EM optimization in a multi-layer perceptron for vector data and the EM optimization along with the consistency optimization or any representation learning technique in a convolutional neural network for image data. The implementation is straightforward compared to many deep approaches. 

\section{Related Work}
\label{se:related_work}
Clustering a set of data points into several categories is generally done by learning a cluster representation, which acts as a basis to cluster the data points into homogeneous groups. The cluster representation is learned by following different approaches such as connectivity based \citep{johnson1967hierarchical}, centroid based \citep{macqueen1967some}, distribution based \citep{mclachlan2019finite}, density modelling based \citep{ester1996density} and subspace-based \citep{hartigan1972direct} methods. Mixture models fall under the distribution-based category, where a statistical distribution represents each cluster. GMM \citep{bishop2006pattern}, in particular, maintains a Gaussian distribution for each cluster and updates those distributions and member assignment iteratively using the EM optimization \citep{em}. k-means \citep{macqueen1967some} can also be thought of as a special case of GMM where the clusters are represented with untilted spheres. Since these traditional clustering algorithms are limited to hand-designed parameters hence limited representability, there have been efforts to incorporate neural networks to model advanced clusters \citep{i2c, adc, deepcluster, dac, wu2019dccm, han2020mitigating}. 

Notably, simple k-means is still being used in many deep clustering techniques \citep{ dec,  deepcluster, peng2018k, kilinc2018learning, tao2020clustering}. These work either synthesize k-means using NNs \citep{peng2018k}, or use k-means in latent abstract space \citep{dec, deepcluster, kilinc2018learning}. Generally, the mapping to latent space is learned via a representation learning method such as autoencoding \citep{dec} or self-supervised methods \citep{kilinc2018learning}. k-means is used to cluster the latent abstract vectors directly \citep{dec} or to generate pseudo-labels to train the network \citep{deepcluster}. In contrast to these works, we let the neural network learn a direct mapping from the input space to class-assignment probabilities while modeling rich cluster distributions following the abstract intuition of mixture modeling.

Although synthesizing k-means using a neural network has been studied before \citep{peng2018k}, realizing mixture modeling and EM for clustering with neural networks is rarely explored as per our knowledge. Neural Expectation-Maximization (N-EM) \citep{greff2017neural} introduces a differentiable clustering method utilizing the EM algorithm. N-EM's objective is to learn the perceptual grouping of a given input by separately identifying the different conceptual entities in the input. N-EM uses neural networks to predict the statistical parameters of cluster distributions (e.g., a single probability for Bernoulli, mean, and variance for Gaussian) from object vectors. These statistical parameters are used to compute the corresponding cluster distribution manually, which is used in EM optimization, and the backpropagation updates the object vector and network parameters. In contrast, we intend to directly approximate the cluster distribution outputs from the final layer neurons of a network for the input data point to reflect the likelihoods of that data point in the corresponding clusters. Our solution is a better replacement for standard clustering objectives such as k-means or GMM.

A successful deep image clustering method should learn general semantic information, which is essential for identifying abstract groups. Therefore, it is vital to enable a neural network to harvest rich features while optimizing the clustering objective. We identify two basic ways to enable such feature learning in the literature. The first one is learning a prior task that enables the network to extract important features \citep{diallo2021deep, dec, kilinc2018learning,  tao2020clustering, van2020scan, tsai2020mice}. The clustering is conducted in the learned abstract space partially or fully using standard algorithms \citep{dec, kilinc2018learning, tao2020clustering} or other clustering objectives as loss functions to train the network \citep{diallo2021deep, han2020mitigating, van2020scan, tsai2020mice}. Learning a self-supervised pre-text task \citep{diallo2021deep, kilinc2018learning, tao2020clustering, van2020scan, tsai2020mice} often enables learning a rich semantic representation that acts as a solid prior for the clustering. Deep Embedded Clustering Based on Contractive Autoencoder (DECCA) \citep{diallo2021deep} uses a two-phase approach of unsupervised feature learning and clustering the learned latent space. The unsupervised feature learning consists of reconstruction of input with a Contractive Autoencoder \citep{rifai2011contractive} and maintaining consistent model response to original and self-augmented inputs. The learned latent space is clustered during the second stage by minimizing the cumulative divergence between embeddings and cluster centers. Semantic Clustering by Adopting Nearest Neighbors (SCAN) \citep{van2020scan}, among its multiple stages, first learns a pre-text task such as instance discrimination \citep{wu2018instance} with a NN. Then, it uses the learned embedding to identify semantically similar neighbors that mostly fall into the same class. Nevertheless, the performance gain of SCAN is mostly attributed to the pre-text task, which acts as a prior to clustering. These multi-stage frameworks often also use the third step to refine the clustered space further using self-labeled supervision \citep{van2020scan, park2021ruc}. In contrast, our two-fold optimization performs feature learning and clustering concurrently, and our clustering optimization is a novel neural mixture modeling method.

The second one is learning the general feature extraction and the clustering simultaneously as a single-stage task.  These methods usually consists of heavy augmentation \citep{i2c, wu2019dccm}, multiple complex losses \citep{adc}, or series of sub-steps \citep{dac, wu2019dccm}. Invariant information clustering (IIC) \citep{i2c}, following the intuition of the consistency optimization, learns a latent mapping from the input by maximizing the mutual information (MI) between the mappings of original and transformed images. MI maximization requires large batch sizes \citep{scae} and repeated sampling where each image is transformed multiple times. DCCM \citep{wu2019dccm} builds on the same consistency optimization and uses pseudo-label and pseudo-graph supervision alongside triplet mutual information optimization. Associative Deep Clustering (ADC) \citep{adc} embeds an image and its transformed version through a CNN and jointly trains the network end-to-end along with another set of centroids. ADC minimizes a sum of multiple losses and requires additional hyper-parameter tuning of the loss weightings. Deep Adaptive Clustering (DAC) \citep{dac} recasts the clustering problem as a binary pairwise classification task. DAC generates the labels by leveraging the learned feature vectors for which constitutes a series of sequential stages.

In contrast to these work,  we propose mixture modeling with a neural network posing as the cluster distribution estimator and the network parameters posing as the cluster distribution parameters. Our framework is a single-stage end-to-end clustering framework. The proposed mixture-EM optimization is a better alternative for k-means and other standard clustering algorithms. In addition, our formulation can be used as the clustering objective along with other initialization methods such as pre-text task learning, consistency optimization, self-label based fine-tuning methods to build rather complex end-to-end multi-stage clustering solutions. This paper discovers only the fusion of consistency optimization to the mixture-EM optimization to cluster image data, resulting in a two-fold single-stage training process.


\section{EM Algorithm for Mixture Model Clustering}
\label{se:em}
Expectation-Maximization \citep{em} is often used to approximate solutions to the maximum likelihood estimate (MLE) and maximum a posterior (MAP) estimate. In mixture modeling, a set of distributions are fitted to the observation space such that the total likelihood is maximized. Due to the difficulties in maximizing this likelihood, EM is often used to optimize an alternative lower bound. Here, we briefly discuss the use of EM to approximate MLE in fitting a mixture of distributions to a given dataset to provide background for our formulation and training approach. 

Let $x$ be an observation of the continuous random variable $X$ in the space $\mathcal{D}$, and $D$ be a set containing $N$ such observations sampled from $\mathcal{D}$ ($D \subset \mathcal{D}$). Let $\theta$ be the parameters which define the clusters. We need to cluster the space $\mathcal{D}$ to $K$ clusters. We introduce a latent discrete variable $Z$ whose outcome is the cluster assignment ($z \in [1\dots K]$). The MLE objective is to find $\theta$ that maximizes the total likelihood marginalized over $Z$: $\mathcal{L}(\theta ; D) = \prod_{x \in D} \sum_{z \in [1\dots K]}  f(x,z \mid \theta)$. Here, $f(x,z  \mid  \theta)$ is the joint probability density of $x$ and $z$, given the parameters $\theta$, viewed as the likelihood of $\theta$ for observed $x$ and $z$: $l(\theta; x,z)$. Note that $f$ denotes a continuous probability whereas $p$ denoted a discrete probability. 

The EM Algorithm alternates between two steps, the E-step and the M-step, to optimize a lower bound to the aforementioned total likelihood. During the $t^{th}$ iteration, in the E-step, the algorithm computes the posteriors for the current $\theta$: $p(z \mid x,\theta^{t})$ and formulates the EM objective:
\begin{equation}
\label{eq:em_objective2}
   Q(\theta \mid \theta^{t})= \sum_{x \in D}  \sum_{z\in [1\dots K]}  p(z \mid x,\theta^{t})  \log [f(x,z  \mid  \theta)] ,
\end{equation}
where $p(z \mid x,\theta^{t})$ is the posterior probability of $Z=z$ given $x$ and current parameters $\theta^{t}$.  
This objective is then optimized w.r.t. $\theta$ in the M-step,
\begin{equation}
\label{eq:m_step}
   \theta^{t+1}    = \argmax_{\theta} Q(\theta \mid \theta^{t}).
\end{equation}
 
At the time of formulation (E-step), $\theta$ carries same values as $\theta^{t}$ as both represent the current parameters. However during the M-step optimization, $\theta$ (in the log term in Eq.~\eqref{eq:em_objective2}) gets updated to new values, i.e., $p(z \mid x,\theta^{t})$ in Eq.~\eqref{eq:em_objective2} is a constant in the M-step optmization.

\section{Towards Formulating Mixture-EM on a Neural Network}
\label{se:method}

From this section onwards, we explain the proposed realization of mixture modeling with EM algorithm using a NN to cluster a dataset end-to-end. Rather than maintaining a set of centroids or Gaussian distributions, we let the NN parameters freely model the clusters with advanced distributions. While the parameters in all layers except the final layer learn shared cluster representations, the parameters in the final-fully connected layer learn cluster-specific representations. We perform EM optimization in an online fashion with batch-wise backpropagation. During the forward pass of the NN, we feed a batch of observations and perform one EM iteration on this batch, where the EM-based loss is calculated and back-propagated. Unlike traditional EM, which fully optimizes the current EM loss in the M-step (Eq.~\eqref{eq:m_step}), we perform only a single step in optimizing the EM loss for the current batch (a gradient descent step). For the next iteration, we feed the next batch and calculate the EM loss again. Such batch-wise EM iteration prevents taking all observations for a single iteration which is inefficient.

First, we approximate $K$ cluster distributions from $K$ output nodes of the NN. To this end, it is vital for all $K$ distributions to behave as probability density functions of input $x$, which implies the $K$ approximated distributions should be continuous, positive, and have a common integral over the sample space. The $K$ distributions must share a common integral; otherwise, a single cluster distribution can easily grow over other distributions, capturing all datapoints. This leads to the trivial solution of all datapoints being allocated to one cluster. We enforce the final layer $K$ neurons to show the PDF behavior by restricting their sigmoid output to the most linear region of the sigmoid with its input normalized over the batch. We then use the approximated cluster distribution outputs to derive the EM objective as in Eq.~\eqref{eq:em_objective2}. The posterior class memberships are approximated for a given input by taking the softmax of the normalized final layer neurons and cluster likelihoods by cluster distribution outputs. Finally, we integrate consistency optimization between original and augmented images into the image clustering to encourage the NN to learn semantically important features. 

Let $g$ be a parametric computation, the NN which learns the cluster characteristics and the cluster assignments. The learnable parameters of the NN now represent the cluster parameters $\theta$. The final layer has $K$ number of nodes, each representing the \emph{relevance} to the respective cluster. Given a batch of $n$ observations $x_{i,i=1,\dots, n}$, for each observation $x_i$, the network outputs $K$ relevance scores $a_i$: $a_i = [a_{i1},\dots, a_{iK}] = g_{\theta}(x_i)$. Here, $a_{ij}$ is the relevance of $x_i$ to $j^{th}$ cluster, which is expected to rise with high degree of membership to cluster $j$. This relevance can further be shown as the output of the composite functions $g_{\theta_s}$ and $g_{\theta_j}$: $a_{ij} = g_{\theta_s,\theta_j}(x_i) = g_{\theta_j}(g_{\theta_s}(x_i))$. If the NN contains $L$ layers, $g_{\theta_s}$ represents the network up-to layer $L-1$, where the parameters $\theta_s$ are shared among all clusters. $g_{\theta_j}$ denotes the mapping from the layer $L-1$ output to the $j^{th}$ neuron of the final layer corresponding to cluster $j$. The parameters $\theta_j$ are exclusive for the cluster $j$.

\section{Approximate Cluster Distributions}
\label{se:cluster_distri}

Now, we move to the most important approximation of this paper, the estimation of the distribution function $h_\theta$ for each cluster, i.e., the probability density of a particular observation $x$ when the cluster assignment $z$ and $\theta$ are known: $f(x \mid z,\theta)$. Let $h_{\theta_s,\theta_j}(x)$ denote the distribution function of the $j^{th}$ cluster: $f(x  \mid Z=j, \theta) = h_{\theta_s,\theta_j}(x)$. Essentially, $h_{\theta_s,\theta_j}(x)$ is a probability density function (PDF) of the observation $x$. Our goal is to derive $h_{\theta_s,\theta_j}(x)$ from the relevance score $g_{\theta_s,\theta_j}(x)$ computed for each batch. To this end, we list the required qualities for the cluster distribution function $h_{\theta_s,\theta_j}$. 
1) $h_{\theta_s,\theta_j}(x)$ should be a continuous function of the observation $x$.
2) $h_{\theta_s,\theta_j}(x)$ should be non-negative for all observations which is particularly important when taking the logarithm. 
3) the integral of $h_{\theta_s,\theta_j}$ over all possible $x$ in the space $\mathcal{D}$ should be 1. 
However, the integral of $h_{\theta_s,\theta_j}$ being a constant that is common to all the clusters is sufficient since dividing the distributions by this constant results in PDFs, and we can neglect this constant in the optimization process. Such an integral restriction regulates the cluster shapes and prevents certain clusters from overpowering other clusters (trivial solutions).

The output of the final layer neuron $j$: $a_{ij} = g_{\theta_s,\theta_j}(x_i)$ is not yet suitable to approximate the distribution of the cluster $j$, although it is related to the degree of the membership of $x_i$ to cluster $j$. Therefore, we transform this relevance to a form which can represent the cluster distribution by another transformation $H$: $ h_{\theta_s,\theta_j}(x) = H(g_{\theta_s,\theta_j}(x))$. We derive this transformation step by step to meet the criterion mentioned above. First, $g_{\theta_s,\theta_j}(x_i)$ is already a continuous function of $x_i$. However, $g_{\theta_s,\theta_j}(x_i)$ (or $a_{ij}$) is the output of a final layer neuron of the NN before any activation, therefore can be negative. The $\mathrm{sigmoid}$ of $a_{ij}$ will transform it to a non-negative value. In addition, the $\mathrm{sigmoid}$ function consists of a nearly linear region which is important as discussed next. 

However, the third criterion is not yet satisfied as the integral of $\mathrm{sigmoid}(a_{ij})$ over all $x \in \mathcal{D}$ does not evaluate to a fixed value common to all the clusters. We enable this property by imposing a common restriction to regulate all cluster distribution integrals within the space $\mathcal{D}$. First, we normalize the input to the $\mathrm{sigmoid}$, the relevance $a_{ij}$ over all such relevance scores to cluster $j$ in the batch.
\begin{equation}
      a^\ast_{ij} = \frac{a_{ij} - \mu_j }{\sigma_j}, \quad where \quad  \mu_j = \frac{1}{n} \sum_{i=1}^n a_{ij} \quad and \quad \sigma^2_j = \frac{1}{n} \sum_{i=1}^n (a_{ij} - \mu_j)^2.
\end{equation}
Here, $a^\ast_{ij}$ is the normalized relevance score of $x_i$ to cluster $j$, where $\mu_j$ and $\sigma_j$ are the mean and the standard deviation of $a_{ij}$ in the batch. Then, we further divide $a^\ast_{ij}$ by another constant $\gamma$ ($\gamma>1$). Thus, the distribution function of cluster $j$ satisfying all criterion is,
\begin{equation}
      h_{\theta_s,\theta_j}(x_i) = \mathrm{sigmoid}(g^\ast_{\theta_s,\theta_j}(x_i)/\gamma) = \mathrm{sigmoid}(a^\ast_{ij}/\gamma).
\end{equation}

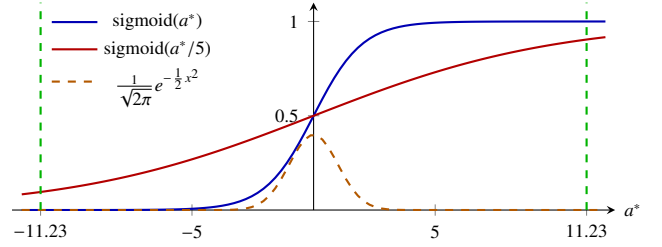
\begin{figure}[t]
	\begin{center}
    \input{sigmoids}
	\end{center}
	\vspace{-0.2in}
	\caption{$\mathrm{sigmoid}$ function and the standard normal distribution. If we assume that the input of the $\mathrm{sigmoid}$ is normalized with zero mean and unit variance, then it is restricted to a limited interval around 0 with high confidence. Even without this assumption, since $a^\ast$ is normalized along a batch-size of 128, it is bounded to [-11.23,11.23] (green vertical lines). We divide $a^\ast$ by $\gamma=5$ to ensure $\mathrm{sigmoid}$ output stays in its mostly linear region for this interval for 128 batch-size. The $\mathrm{sigmoid}$ function, with its zero mean linear region input, acts as a PDF of $x$.}
	\label{fig:sigmoid}
	\vspace{-0.1in}
\end{figure}

We now explain the motivation and the justification for such normalization of relevance score $a_{ij}$ and division by $\gamma$. Referring to Fig.~\ref{fig:sigmoid}, if we assume that the input to the $\mathrm{sigmoid}$ is a standard normal variable with zero mean and unit variance, the input of the $\mathrm{sigmoid}$ activation function is restricted to a small interval around zero as larger inputs are unlikely (99.9\% confidence interval of standard normal score is $[-3.29,3.29]$). Even without this assumption, since we normalize a set of $n$ points $[a_{1j},\dots, a_{nj}]$, the normalized values are bounded within the interval of $\left[ \frac{-(n-1)}{\sqrt{n}}, \frac{(n-1)}{\sqrt{n}} \right]$ \citep{shiffler1988maximum}. For example, if we normalize $a_{ij}$ over a batch of 128, for any observation $x_i$, the normalized relevance score $a^\ast_{ij}$ is bounded within the interval $[-11.23,11.23]$. Furthermore, to keep the $\mathrm{sigmoid}$ output within its mostly linear region for all $a^\ast_{ij}$ which falls within this interval, we divide $a^\ast_{ij}$ by $\gamma$ ($\gamma>1$). $\gamma$ is dependent on the batch-size. If the batch-size is 128, we set $\gamma$ to 5 to make sure the $\mathrm{sigmoid}$ output is within the mostly linear region, as depicted in Fig.~\ref{fig:sigmoid}. 

Normalization of relevance score $a_{ij}$ over the batch ($a^*_{ij}$) and further dividing by $\gamma$ ($\gamma>1$) makes couple of important restrictions to the cluster distribution $h_{\theta_s, \theta_j}(x_i) = \mathrm{sigmoid}(a^*_{ij}/\gamma)$ as follows;
\begin{enumerate}
    \item The average input to the $\mathrm{sigmoid}$ ($a^*_{ij}/\gamma$ or $g^*_{\theta_s, \theta_j}(x_i)/\gamma$) over the batch is zero:
    \begin{equation}
         \frac{1}{n}\sum_{i=1}^n a^*_{ij}/\gamma = 0.
        \label{eq:mean_sig}
    \end{equation}
    
    \item For any observation $x_i$, $\mathrm{sigmoid}(a^*_{ij}/\gamma)$ limits to its mostly linear region. In addition, since the average input to the $\mathrm{sigmoid}$ is zero, we can approximate the average of $\mathrm{sigmoid}(a^*_{ij}/\gamma)$ over the batch by 0.5:
    \begin{equation}
        \frac{1}{n}\sum_{i=1}^n h_{\theta_s, \theta_j}(x_i) =  \frac{1}{n}\sum_{i=1}^n \mathrm{sigmoid}(a^*_{ij}/\gamma) = 0.5.
        \label{eq:e_sig_x}
    \end{equation}
\end{enumerate}
With these conditions, using the Monte Carlo integral estimation \citep{metropolis1949monte}, we can show that over all $x \in \mathcal{D}$, cluster $j$ distribution $h_{\theta_s, \theta_j}$ integrates to a value which is common for all clusters. Let us denote the integral of cluster $j$ distribution $h_{\theta_s, \theta_j}(x)$ over the space $\mathcal{D}$ by $I_j$:
\begin{equation}
    I_j = \int_{\mathcal{D}}  h_{\theta_s,\theta_j}(x) \,dx.
\end{equation}
Here, $x$ is an $m$-dimensional observation from the space $\mathcal{D}$ ($\mathcal{D} \subset \mathrm{R}^m$). The objective is to approximate this integral by uniformly sampled batch of $n$ observations $x_{i \mid i=1,\dots, n}$. Given the batch of $n$ uniform samples in the space $\mathcal{D}$, the Monte Carlo method approximate $I_j$ by,
\begin{equation}
    I_j \approx V \frac{1}{n} \sum_{i=1}^{n} h_{\theta_s,\theta_j}(x_i),
    \label{eq:monte}
\end{equation}
where V is the volume of the $m$-dimensional space $\mathcal{D}$: $ V = \int_{\mathcal{D}}  \,dx$.
The average of distribution outputs over $n$ samples is 0.5 as per Eq.~\eqref{eq:e_sig_x}. Thus, the Monte Carlo estimation for the integral of the cluster $j$ distribution over the space $\mathcal{D}$ ($I_j$) becomes,
\begin{equation}
    I_j \approx  0.5V.
\end{equation}
The approximated integral for cluster $j$ distribution ($0.5V$) is common for all cluster distributions $h_{\theta_s, \theta_j}(x)|_{j=1,\dots, K}$. Therefore, within the sample space $\mathcal{D}$, these distributions act as PDFs with a common integral. This integral regularization prevents certain clusters overpowering other clusters, i.e., trivial solutions or empty clusters.

\section{Deploying EM Batch-Wise}
\label{se:deploy_em}

\begin{figure*}[t]
	\begin{center}
	\includegraphics[width=\linewidth]{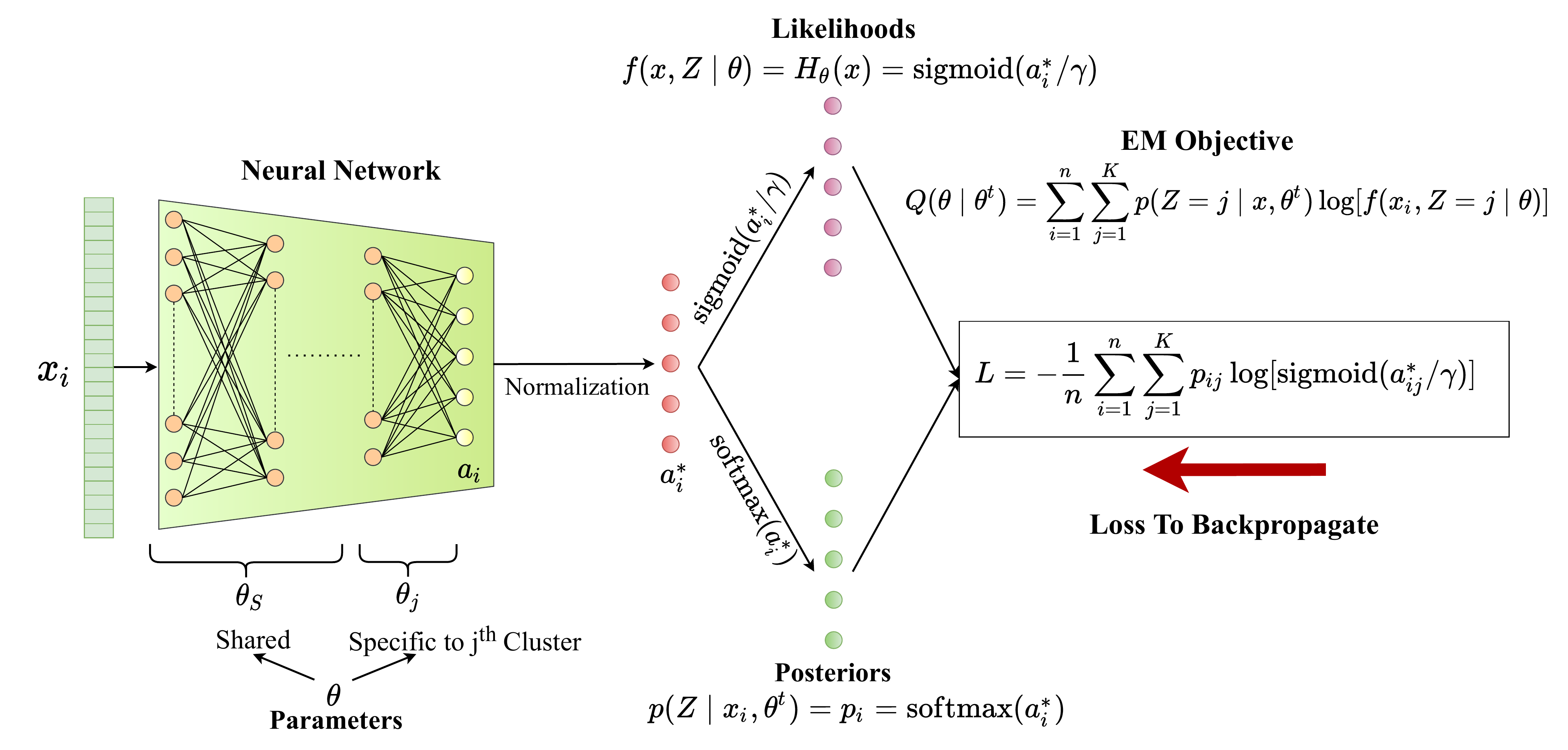}
	\end{center}
	\vspace{-0.2in}
	\caption{Adopting Mixture-EM to train a NN end-to-end for clustering. The NN final layer relevance scores are normalized and used to calculate cluster likelihoods. Same normalized relevance scores are used to calculate posteriors. Using the likelihoods and posteriors, the EM-based loss is derived which is backpropagated.}
	\label{fig:clusternet}
	\vspace{-0.1in}
\end{figure*}

With the cluster distributions defined, we move to the batch-wise EM iterations. During the $t^{th}$ forward pass for a given batch $x_{i, i=1,\dots, n}$, first, we calculate the E-step posterior probabilities $p(z \mid x,\theta^t)$. Given the input $x_i$ and parameters $\theta$, the probability of the assigned cluster being $j$ is,
\vspace{-0.08in}
\begin{equation}
    p(Z=j \mid x_i,\theta) = \frac{ p(Z=j \mid \theta)f(x_i \mid Z=j,\theta) }{\sum_{k=1}^K p(Z=k \mid \theta)f(x_i \mid Z=k,\theta)},
\end{equation}
where $p(z \mid \theta)$ is the prior of cluster assignment. These priors are calculated from the posteriors of the previous step. However, since we experiment in evenly divided datasets, we assign the priors of all clusters to $1/K$. This simplifies the posterior to the normalized cluster distribution:
\vspace{-0.08in}
\begin{equation}
    p(Z=j \mid x_i,\theta) = \frac{h_{\theta_s,\theta_j}(x_i)}{\sum_{k=1}^K h_{\theta_s,\theta_k}(x_i)}.
\end{equation}
However, considering the ease of numerical optimization, we simplify this ratio between $\mathrm{sigmoid}$ values to the ratio between un-normalized exponentials, hence the $\mathrm{softmax}$ function. Therefore, we approximate the posterior $p(Z=j \mid x_i,\theta^t)$ by the $\mathrm{softmax}$ activated $a^\ast_{ij}$ (Denoted by $p_{ij}$ later):
\vspace{-0.08in}
\begin{equation}
    p(Z=j \mid x_i,\theta^t) \: \approx \:  \frac{e^{a^\ast_{ij}}}{\sum_{k=1}^{K}e^{a^\ast_{ik}}} = p_{ij}.
\end{equation}
Then, we estimate the joint probability density of observation $x$ and cluster assignment $z$ given parameters $\theta$: $f(x,z  \mid  \theta)$, i.e., the likelihood of $\theta$ for observed $x$ and $z$: $l(\theta; x,z)$. This can be expanded as $p(z  \mid  \theta)f(x  \mid z, \theta)$. As we set the prior for cluster assignment $p(z  \mid  \theta)$ to constant $1/K$, it can be disregarded in the optimization. Therefore, the joint density can directly be approximated by the conditional density $f(x  \mid z, \theta)$ hence the cluster distribution is,
\vspace{-0.08in}
\begin{equation}
     f(x,z  \mid  \theta) = h_{\theta_s,\theta_j}(x_i) =  \mathrm{sigmoid}(a^\ast_{ij}/\gamma).
\end{equation}
Once the E-step posteriors and joint probability densities (likelihoods of $\theta$) are formulated for the batch, we compute the EM-based loss function which corresponds to the EM objective in Eq.~\eqref{eq:em_objective2} by,
\vspace{-0.08in}
\begin{equation}
\label{eq:em_loss}
   L_{\theta \mid \theta^t}= -\frac{1}{n} \sum_{i = 1}^n  \sum_{j=1}^K  p_{ij}  \log [\mathrm{sigmoid}(a^\ast_{ij}/\gamma)]. 
   \vspace{-0.08in}
\end{equation}
Keeping $p_{ij}$ constant, we backpropagate this loss and update parameters $\theta$ as the objective is to maximize the total likelihood subjected to the current posterior probabilities. Fig. \ref{fig:clusternet} summarizes this formulation.

The proposed method approximates the sample space with a batch of samples at each iteration. It is vital to normalize the relevance scores and divide by $\gamma$ to restrict them to the most linear region of the sigmoid activation to obtain the PDF behavior. The value of $\gamma$ depends on the selected batch size. For our experiments, we used a batch size of 128. Therefore, the normalized relevance scores fall between [-11.23, 11.23], highly scattered around zero. To ensure the sigmoid of these scores are within its most linear region, we maintain $\gamma=5$. If we use a different batch size, we have to tune $\gamma$ so that the input to the sigmoid lies within its most linear region. In addition, since the proposed framework performs EM iterations batch-wise, the EM optimization sees a batch of samples that represents the entire sample space in each iteration. Therefore, the larger the batch size, the batch can better represent the sample space for each EM iteration. Meanwhile, having smaller batches helps regulate the optimization process by adding more noise to the sample space approximation.
\section{Fusing with Consistency Optimization for Image Clustering}
\label{se:aug}

To cluster high dimensional data such as images, training a NN with a pure clustering objective only on original data is insufficient. The network could easily get overfitted to the lower level textures and patterns which are unnecessary for the clustering task. Therefore, the network needs to extract general abstract features (e.g., body patterns, poses) from images relevant to capturing the class. To enable such rich feature extraction alongside mixture-EM optimization, we incorporate transformed images into the learning process. The transformation $T$ converts an original image $x_i$ to its transformed version $x^{tr}_i$: $x^{tr}_i = T(x_i)$. It consists of basic data augmentation such as random crop, shift, rotation, scale, and random adjustment of image brightness, contrast, saturation, and hue. We compute the relevance scores for the transformed images as before: $a^{tr}_{i} = g_{\theta}(x^{tr}_i)$. We add another term to the EM loss in Eq.~\eqref{eq:em_loss}, the log-likelihood for the transformed image weighted by the posterior of the original image. Thus, the loss becomes,
\vspace{-0.08in}
\begin{equation}
\label{eq:em_loss_aug}
   L_{\theta \mid \theta^t} = -\frac{1}{n} \sum_{i = 1}^n  \sum_{j=1}^K   p_{ij} \left [ \log [\sigma(a^\ast_{ij}/\gamma)] + \log [\sigma(a^{tr*}_{ij}/\gamma)] \right].
\end{equation}
Optimizing the log-likelihoods of both original and transformed images in favor of the posterior for the original image ($p_{ij}$) encourages the network to maintain similar behavior for both original and transformed images. Further continuing on such motivation, we use a concept similar to consistency regularization \citep{laine2016temporal} to encourage the network to maintain similar outputs for the original and transformed images. Once the model outputs posterior probabilities $q_i$ to the transformed input $x^{tr}_i$:  $ q_i = \mathrm{softmax}(g^\ast_{\theta}(x^{tr}_i))$, we minimize the KL divergence between the posteriors $p_i$ and $q_i$ for the original image and the transformed image respectively,
\vspace{-0.08in}
\begin{equation}
\label{eq:kl_loss}
    D_{KL} (p_i  \mid  \mid  q_i) = \sum_{j=1}^{K} p_{ij} \mathrm{log}  \frac{p_{ij}}{q_{ij}}. 
    \vspace{-0.08in}
\end{equation}
We keep the posteriors of the original image $p_i$ constant in optimizing the KL divergence, making them temporary soft labels for the augmented image response $q_i$. We embed this optimization into the main EM optimization by performing gradient steps in optimizing both objectives (Eq.~\eqref{eq:em_loss_aug} \& Eq.~\eqref{eq:kl_loss}) one after the other for each batch, using two separate optimizers. This leads to a two-fold optimization process as shown in Algorithm \ref{alg:em}. 
\begin{algorithm}[t]
	\caption{Two-fold optimization for a given batch $[x_{i,i=1\dots n}]$, its transformed batch $[x^{tr}_{i,i=1,\dots, n}]$ and the NN which is parameterized by $\theta$}
	\label{alg:em}
	\begin{algorithmic}
		\STATE {\bfseries EM Optimization}
		\STATE Compute cluster relevances scores
		\STATE $a_i = [a_{i1} ,\dots, a_{ik}] = g_{\theta}(x_i)$. Similarly \ $a^{tr}_i = g_{\theta}(x^{tr}_i)$
		\STATE Normalize relevances $a^\ast_{ij} = \frac{a_{ij} - \mu_j }{\sigma_j}$. Similarly $a^{tr*}_{ij}$.
		\STATE Compute posterior probabilities $p_i = \mathrm{softmax} (a^\ast_i)$.
		\STATE Compute clustering loss: $L_{\theta \mid \theta^t},$ 
		\STATE $-\frac{1}{n} \sum_{i = 1}^n  \sum_{j=1}^K  p_{ij} \left [  \log [\sigma(a^\ast_{ij}/\gamma)] + \log [\sigma(a^{tr*}_{ij}/\gamma)] \right]$ (Eq.~\eqref{eq:em_loss_aug})
		\STATE Backpropagate loss and update parameters $\theta$   
		\STATE {\bfseries Consistency Optimization for Augmented Images}
		\STATE Compute posteriors $p_i$ with updated $\theta$
		\STATE $p_i = \mathrm{softmax} (g^\ast_{\theta}(x_i))$
		\STATE Compute posteriors for augmented images
		\STATE $q_i = \mathrm{softmax} (g^\ast_{\theta}(x^{tr}_i))$
		\STATE Step in optimize the KL Divergence ($p_{ij}$ constant)
		\STATE $D_{KL} (p_i  \mid  \mid  q_i) = \sum_{j=1}^{K} p_{ij} \mathrm{log}  \frac{p_{ij}}{q_{ij}} $ (Eq.~\eqref{eq:kl_loss})
	\end{algorithmic}	
\end{algorithm}

The posterior for original images $p_i$ updates to a better posterior after one gradient step of EM optimization as the network parameters get updated in favor of the current posteriors. The KL objective intends to encourage the network to maintain the current network response for the original image $p_i$, for its augmented image. Optimizing these two objectives together is inefficient. Because when the posteriors for the original images $p_i$ gets updated to better values ($p_i^{new}$), the KL objective forces the posteriors for the augmented images $q_i$ to stay closer to the current (old) $p_i$. Therefore we optimize these two objectives alternatively. After one EM optimization step, to optimize the KL objective, we freshly calculate the posteriors for the original image $p_i$ since the network has now been updated.

\section{Time Complexity Analysis}
\label{se:complexity}
Considering the dimension of each sample, the number of samples to cluster, the number of clusters (i.e., number of final layer neurons), and the number of iterations for convergence, the proposed optimization of NN carries a similar time complexity to k-means or GMM. Let us consider a task of clustering $n$ samples to $K$ clusters, with each sample being an image of dimension $d=M\times N$. If k-means takes $i_{k-means}$ number of iterations over set $D$ for convergence, the time complexity of clustering this space with k-means is $\mathcal{O}(ndKi_{k-means})$. 

Let us assume we employ a convolutional network of $N_c$ convolutional layers followed by a fully-connected layer for the proposed clustering. Let every convolutional layer contains $f$ number of $w\times w$ filters without any down-sampling, and the fully connected layer outputs $K$ nodes. The running time for a sample is therefore $\mathcal{O}(dfw^2)$ for the first convolutional layer, $\mathcal{O}(\sum_{N_{c-1}}df^2w^2) = \mathcal{O}(df^2w^2)$ for the rest of convolutions and $\mathcal{O}(dfK)$ for the fully-connected layer. This is a valid upperbound even for a network with down-sampling layers. Therefore, the time taken for a single image to forward pass through the NN is  $\mathcal{O}(dfw^2 + \sum_{N_{c-1}}df^2w^2 + dfK)$. This can be simplified to $\mathcal{O}(dp)$ where $p$ denotes the total number of parameters of the NN ($p=fw^2 + \sum_{N_{c-1}}f^2w^2 + fK$). Thus, the complexity of our framework considering $n$, $d$, $p$ and $i_{NN}$ is $\mathcal{O}(ndpi_{NN})$. During training, the backward pass through the network is of similar time complexity to the forward pass. If we fix all layers except the final layer, the total running time for one sample becomes $\mathcal{O}(dK)$ depending on the number of final layer neurons $K$. Thus, the time complexity of the proposed clustering for $n$ samples and $i_{NN}$ number of iterations over set $D$ (number of epochs) now becomes $\mathcal{O}(ndKi_{NN})$, which is similar to k-means.

\section{Experiments}
\label{se:exp}
\subsection{2-Dimensional Space}
\label{ss:2d} 

\begin{figure}[ht]
	\begin{center}
	\begin{subfigure}{0.49\columnwidth}
	    \includegraphics[width=\columnwidth]{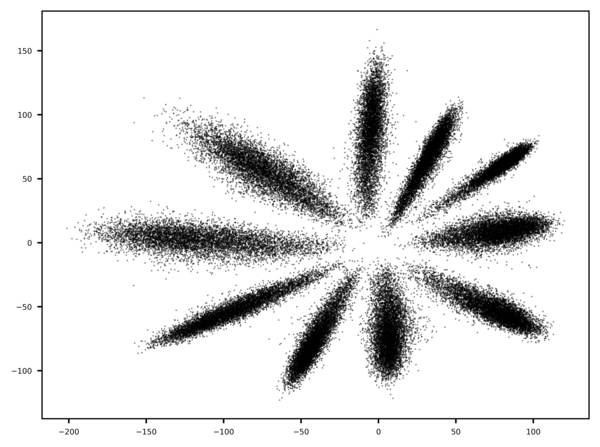}
	    \caption{Input Space}
	    \label{sfig:input_space}
    \end{subfigure}
    \begin{subfigure}{0.49\columnwidth}
	    \includegraphics[width=\columnwidth]{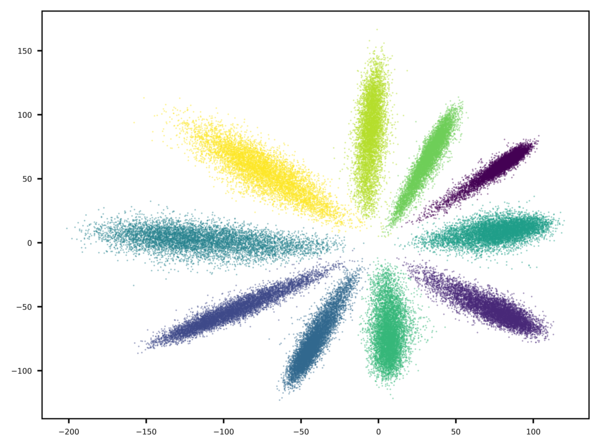}
	    \caption{Categorized Space}
	    \label{sfig:cat_input_space}
    \end{subfigure}
	\end{center}
	\vspace{-0.2in}
	\caption{The 2-d sample space generated from the MNIST dataset and the clustered set after training with our algorithm. The algorithm identifies observable clusters. See Fig.~\ref{fig:distributions} for the cluster distribution plots.}
	\label{fig:sample_spaces}
	\vspace{-0.1in}
\end{figure}

To validate our algorithm and study the cluster distribution behavior, we first conduct a small-scale clustering experiment on a 2-d space created from the MNIST \citep{mnist} dataset. To create 2-d data from MNIST, we train a CNN in MNIST in a supervised manner. The network contains a bottleneck layer of 2 nodes before the final 10-node layer. Once trained, we extract the bottleneck output of the network that contains 70,000 2-d points as shown in Fig.~\ref{sfig:input_space}. We use these data points as the set $D$ for the clustering. This supervised setting for dimensionality reduction enables the 2-d samples to scatter in observable clusters better than unsupervised techniques \citep{wold1987principal, maaten_tsne, kramer1991nonlinear}. 

To cluster this 2-d space, we use a three-layer perceptron with two 32-node hidden layers and a 10-node final layer (10 clusters). We use the mixture-EM optimization as in Eq.~\eqref{eq:em_loss} which uses only original data points. We train the NN with a batch size of 128, setting $\gamma$ to 5, and using an Adam optimizer \citep{kingma2014adam} with a learning rate (LR) of 0.001. The algorithm converges within ten epochs. Fig.~\ref{sfig:cat_input_space} shows the clustered space, where the network identifies the observable clusters. The final layer relevance score normalization does not use the affine transformation as in conventional batch-normalization \citep{bn} since it can reverse the normalization effect. Furthermore, while relevance score normalization can be done dynamically, using a running mean and a running standard deviation smoothens the learning and enables inference with different batch sizes.

\begin{figure}[ht]
	\begin{center}
	\begin{subfigure}{\columnwidth}
	    \begin{subfigure}{0.32\columnwidth}
	        \includegraphics[width=\columnwidth]{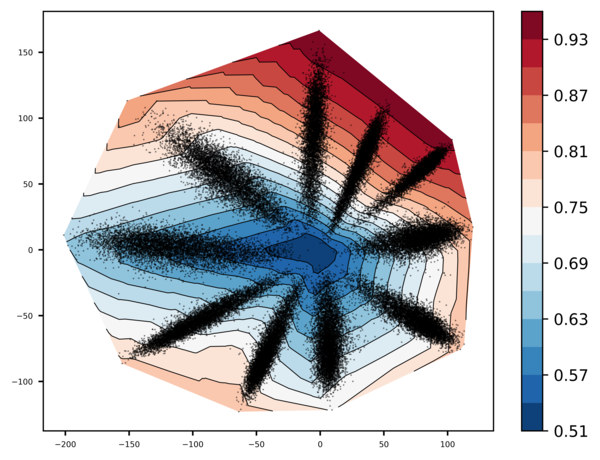}
	        \caption*{cluster 4}
	    \end{subfigure}
	    \begin{subfigure}{0.32\columnwidth}
	        \includegraphics[width=\columnwidth]{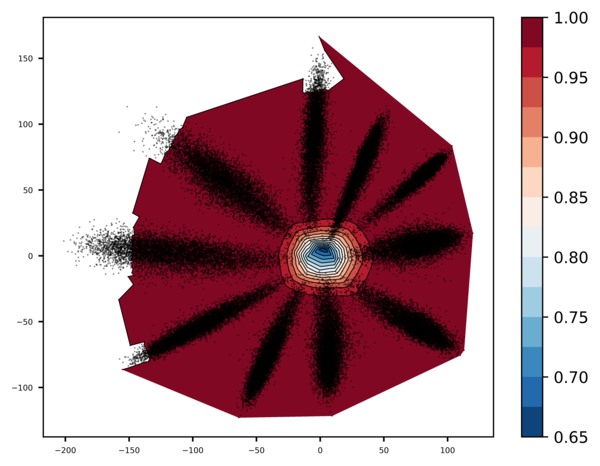}
	        \caption*{cluster 6}
	    \end{subfigure}
	    \begin{subfigure}{0.32\columnwidth}
	        \includegraphics[width=\columnwidth]{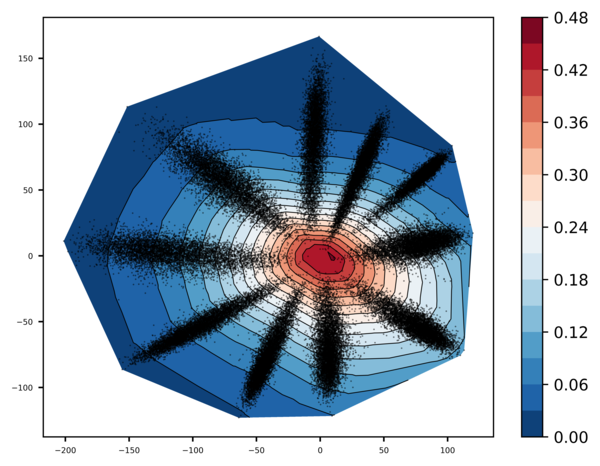}
	        \caption*{cluster 10}
	    \end{subfigure}
	    \vspace{-0.08in}
		\caption{Without normalizing relevance scores}
		\label{sfig:dist_nobn}
	\end{subfigure}
		\begin{subfigure}{\columnwidth}
	    \begin{subfigure}{0.32\columnwidth}
	        \includegraphics[width=\columnwidth]{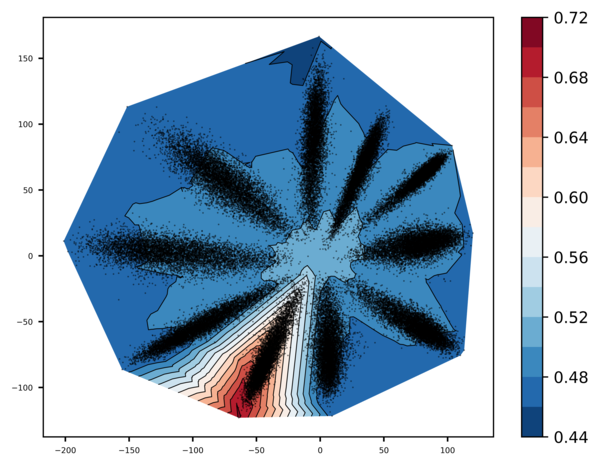}
	        \caption*{cluster 4}
	    \end{subfigure}
	    \begin{subfigure}{0.32\columnwidth}
	        \includegraphics[width=\columnwidth]{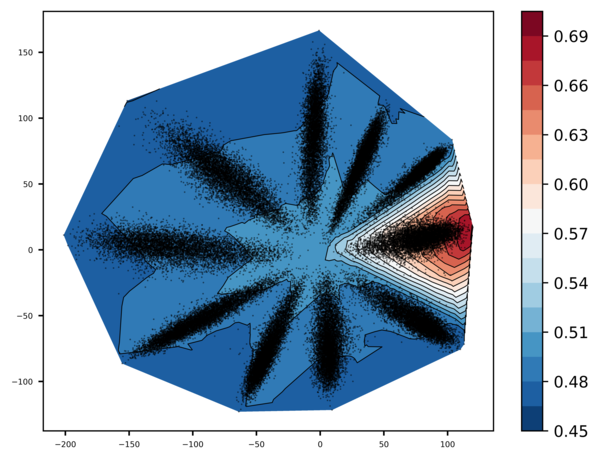}
	        \caption*{cluster 6}
	    \end{subfigure}
	    \begin{subfigure}{0.32\columnwidth}
	        \includegraphics[width=\columnwidth]{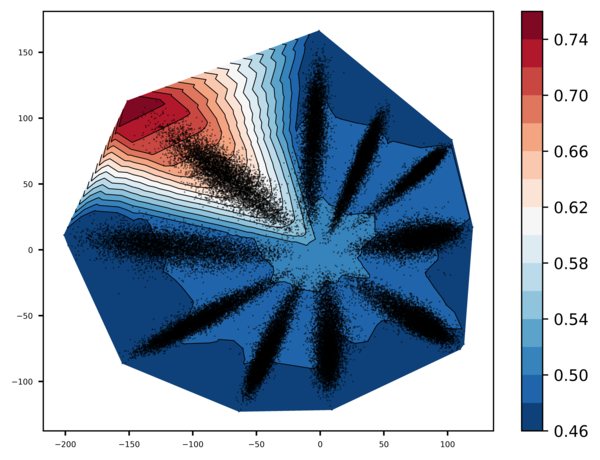}
	        \caption*{cluster 10}
	    \end{subfigure}
	    \vspace{-0.08in}
		\caption{With normalized relevance scores}
		\label{sfig:dist_bn}
	\end{subfigure}
	\end{center}
	\vspace{-0.2in}
	\caption{Importance of relevance score normalization: Cluster distributions of chosen clusters when trained in the 2-d space. (a) Without relevance score normalization, some clusters overpower other clusters by acquiring all members. Here, cluster 6 dominates the sample space by capturing all observations. (b) With relevance score normalization, the cluster distributions act as PDFs with a common integral over the input space. Distributions show high likelihoods for each observable group.}
	\label{fig:distributions}
	\vspace{-0.1in}
\end{figure}

In Fig.~\ref{fig:distributions}, we show the contour plots of the cluster distributions ($f(x  \mid Z=j, \theta)$ or $h_{\theta_s,\theta_j}(x)$) with the observation $x$. We choose three cluster distributions among the 10 clusters corresponding to $4^{th}$, $6^{th}$ and $10^{th}$ nodes of the final layer. Fig.~\ref{sfig:dist_nobn} shows the chosen cluster distributions when trained with un-normalized relevance scores $a_{ij}$ where the distributions are not explaining possible clusters. Also, their integrals over the space do not seem to evaluate to a common value. Cluster 6 distribution shows maximized likelihoods for all data points, where cluster 10 distribution is fairly low for all observations. In this scenario, the model collapses to a trivial solution by assigning all samples to cluster 6. Fig.~\ref{sfig:dist_bn} shows the same cluster distributions  when trained with normalized relevance scores $a^\ast_{ij}$. The contour plots illustrate that each distribution captures an observable cluster of points by showing high likelihoods for those points. These distributions act close to PDFs since they empirically share a common integral over the observation space.  These plots validate that our relevance score normalization is crucial for clustering.

\begin{table}[ht]
    \caption{Average of estimated cluster distribution outputs for the three clusters in Fig.~\ref{fig:distributions} for a given batch of 128 samples. Without relevance score normalization, they reach different values without any restriction. Cluster 6 has the highest average, closer to 1. With relevance score normalization, they all reach 0.5.}
    \label{tab:avg_dist}
    \small
    \vspace{-0.1in}
	\begin{center}
	\begin{tabular}[width=\columnwidth]{@{}l|ccr@{}}
		\toprule
 	    Cluster index $j$ & 4  & 6 & 10  \\
 	    \midrule 
		$\frac{1}{n} \sum_{i=1}^n h_{\theta_s,\theta_j}(x_i)$ with $a_{ij}$s (un-normalized)
		    &0.736 &0.999 &0.108 \\
		\midrule 
		$\frac{1}{n} \sum_{i=1}^n h_{\theta_s,\theta_j}(x_i)$ with $a^\ast_{ij}$s (normalized)
		    &0.498 &0.507 &0.504 \\
		\bottomrule
	\end{tabular}
	\end{center}
	\vspace{-0.2in}
\end{table}

We further estimate the cluster distribution integrals for these 3 clusters experimentally to validate the PDF behavior. As shown in Eq.~\eqref{eq:monte}, the integral of the cluster distribution $h$ over the space $\mathcal{D}$ is proportional to the average output of $h$ over $n$ uniform samples. We also showed that the average cluster distribution output of a batch of $n$ samples reaches 0.5 due to the relevance score normalization and constraint to the sigmoid linear region (Eq.~\eqref{eq:e_sig_x}). To empirically observe this, we calculate the average cluster distribution output $\frac{1}{n} \sum_{i=1}^n h_{\theta_s,\theta_j}(x_i)$ for these 3 clusters over a given batch of $n=128$ samples and report in Table \ref{tab:avg_dist}. Without the relevance score normalization, the average cluster distribution outputs over the batch show different values, and the dominating cluster 6 shows an average close to 1. All three clusters show an average distribution output close to 0.5 for the batch when we use relevance score normalization. Hence, they share a common integral of $0.5V$ ($V$ is the volume of $\mathcal{D}$ as in Eq.~\eqref{eq:monte}) over space $\mathcal{D}$ and act as PDFs of $x$.

\subsection{Clustering Image Datasets}
\label{ss:cluster_image} 

\begin{table}[ht]
	\caption{Network Architectures. C$n$ denotes a convolutional operation with $n$ filters. M stands for Max-pooling. F$n$ denotes a fully-connected layer with $n$ output nodes.}
	\label{tab:net_archi}
	\vspace{-0.2in}
	\begin{center}
	\footnotesize
	\begin{tabular}[width=\columnwidth]{@{}llr@{}}
		\toprule
 	    \textbf{Dataset}        & \textbf{Architecture} & \textbf{Params}   \\
		\midrule 
		MNIST         & C64 M C128 M C256  F32 F10                 & 0.8M            \\
 	    CIFAR        & C64 C64 M C128 C128 M C256 C256 F$n$   & 1.3M\\
 	    STL10         & C64 C64 M C128 C128 M C256 C256             & 2.7M \\
 	                  & M C256 C256 F10  & \\
		\bottomrule
	\end{tabular}
	\end{center}
	\vspace{-0.2in}
\end{table}

We further test our algorithm on four image datasets commonly used for unsupervised clustering, 
STL10 \citep{stl10} \footnote[1]{STL10 - \url{https://cs.stanford.edu/~acoates/stl10/}}, 
CIFAR10/100 \citep{cifar100} \footnote[2]{CIFAR10 \& CIFAR100 - \url{https://www.cs.toronto.edu/~kriz/cifar.html}} and 
MNIST \citep{mnist} \footnote[3]{MNIST - \url{http://yann.lecun.com/exdb/mnist/}}.
STL10 consists of 13k labeled samples and 100k unlabelled samples. We only cluster the labeled set as the unlabelled set contains additional classes. CIFAR100 contains 100 classes and the data are further abstracted to 20 super-classes, each super-class containing 5 classes. Following other clustering work \citep{dec, i2c, dac}, we cluster the CIFAR100 dataset to the 20 super-classes and other datasets to the standard numbers of categories. We use a 9-layer CNN for the STL10 dataset, a 7-layer network for CIFAR10/100, and a 5-layer network for the MNIST dataset. The network architectures are detailed in Table~\ref{tab:net_archi}. Before feeding the images to the NN, if RGB, we convert them to single-channel grayscale. We process the single-channel images with vertical and horizontal Sobel filters. Thus, NN input is a stack of two planes of similar height and width to the original image ($2\times H\times W$), carrying vertical edges and horizontal edges. Such pre-processing prevents the network from overfitting to colors and enables learning of general structures. 

We use the two-fold optimization shown in Sec.~\ref{se:aug} and Algorithm~\ref{alg:em}, maintaining two Adam \citep{kingma2014adam} optimizers, one for EM optimization (LR = 5e-5) and the other for consistency optimization (LR = 1e-4). It is important to assign a higher LR for consistency optimization as if learning general semantic features gets high priority over learning cluster representations; clustering will be richer and more accurate. We train our models for 250 epochs with a batch size of 128 and $\gamma=5$, with the full datasets apart from STL10 where we use the labeled set.

\begin{table*}[!t]
	\caption{Unsupervised classification accuracy (\%) comparison. We cluster CIFAR100 into 20 superclasses. $\dagger$ denotes approaches that get the support of k-means. Our two-fold optimization surpasses all traditional clustering algorithms, all single-stage deep algorithms that still rely on k-means, and even state-of-the-art single-stage approaches in some cases. $\ast$ denotes DeepCluster \citep{deepcluster} and ADC \citep{adc} performance figures produced by Ji \etal, \citep{i2c}.}
	\label{tab:sota}
	\vspace{-0.2in}
	\begin{center}
	\footnotesize
	\begin{tabular}[width=\columnwidth]{@{}lcccccccr@{}}
		\toprule
 	     \textbf{Approach}    &   \multicolumn{2}{c}{\textbf{STL10}} &  \multicolumn{2}{c}{\textbf{CIFAR10}}
 	     &  \multicolumn{2}{c}{\textbf{CIFAR100}} &  \multicolumn{2}{c}{\textbf{MNIST}}   \\
 	      &Acc &NMI  &Acc &NMI &Acc &NMI &Acc &NMI\\
		\midrule 
		 K-means \citep{zelnik2005self}                  &19.2 	&12.5   &22.9  &8.7   &13.0 &8.4  &57.2   &50.0\\
		 Spectral Clustering \citep{wang2014optimized}  &15.9 	&9.8   &24.7  &10.3  &13.6  &9.0  &69.6  &66.3\\
		 JULE \citep{yang2016joint}                     &27.7 	&18.2   &27.2  &19.2  &13.7  &10.3  &96.4  &91.3\\
		 \midrule
		 Triplets \citep{schultz2004learning} $\dagger$          &24.4 &- 	&20.5  &-   &9.94 &-  &52.5 &-\\
		 AE \citep{bengio2007greedy} $\dagger$                   &30.3 &25.0 	&31.4  &23.4  &16.5 &10.0 &81.2 &72.6\\
		 Sparse AE \citep{ng2011sparse} $\dagger$                &32.0  &25.2 	&29.7  &24.7  &15.7 &10.9 &82.7 &75.7\\
		 Denoising AE \citep{vincent2010stacked} $\dagger$       &30.2  &22.4	&29.7  &25.1  &15.1 &11.1 &83.2 &75.6\\
		 Var. Bayes AE \citep{kingma2013auto} $\dagger$          &28.2 &20.0	&29.1  &24.5  &15.2 &10.8 &83.2 &73.6\\
		 SWWAE \citep{zhao2015stacked} $\dagger$                 &27.0 &19.6	&28.4  &23.3  &14.7 &10.3 &82.5 &73.6\\
		 GAN \citep{radford2015unsupervised} $\dagger$           &29.8 &21.0	&31.5  &26.5  &15.1 &12.0 &82.8 &76.4\\
		 DEC \citep{dec} $\dagger$                               &35.9 &27.6	&30.1  &25.7  &18.5 &13.6 &84.3 &77.2\\
		 K-meansNet \citep{peng2018k}                            &-    &-  &20.23   &6.87 &- &-  &87.76  &78.70\\
		 DeepCluster \citep{deepcluster} $\dagger$               &33.4* &-	&37.4* &-   &18.9* &- &65.6* &-\\
	     \midrule
	     DECCA \citep{diallo2021deep}  &- &-    &- &-   &- &-   &96.37 &0.9087  \\
	     SCAE \citep{scae}                 &- 	&-    &33.48  &-   &-  &-     &\textbf{99.0} &-\\
		 DAC \citep{dac}                   &47.0 &36.6	&52.2  &40.0  &23.8  &18.5    &97.8  &93.5\\
		 ADC \citep{adc}                   &53.0 &-	&32.5  &-  &16.0*  &-    &\textbf{99.2} &-\\
		 IIC \citep{i2c}                   &59.8 &49.6	 &\textbf{61.7} &\textbf{51.1}   &25.7 &\textbf{22.5}  &\textbf{99.2} &-\\
		 IIC \citep{i2c} our setting      &47.12 &0.4102	 &44.17 &0.3489   &16.18 &0.0988  &95.72 &0.9396\\
		 
		 \midrule
		 EM Optimization (Eq.~\eqref{eq:em_loss_aug}) &49.61 & 0.4199  &49.53  &0.3959  &19.36 &0.1223  &98.44 &0.9567\\
		 Two-Fold Optimization          &\textbf{63.84} &0.503  &\textbf{57.97}  &0.4703    &\textbf{25.94}  &0.1972 &\textbf{98.88}  &0.9674\\
		                       &\scriptsize{$\pm$ 2.6} & \scriptsize{$\pm$ 0.0213}	&\scriptsize{$\pm$ 3.03} & \scriptsize{$\pm$ 0.0204}    &\scriptsize{$\pm$ 0.8} & \scriptsize{$\pm$ 0.0041}  &\scriptsize{$\pm$ 0.07}  & \scriptsize{$\pm$ 0.0016}\\
		\bottomrule
	\end{tabular}
	\end{center}
	\vspace{-0.2in}
\end{table*}

We compare our approach with traditional and state-of-the-art deep clustering methods in Table~\ref{tab:sota}. Here, we evaluate the trained model's unsupervised classification accuracy and the normalized mutual information (NMI). Since we cannot expect the predicted class indexes to match the labels due to the unsupervised training setting, we use the Hungarian algorithm \citep{kuhn1955hungarian} to assign the predicted cluster index to the actual label as a linear sum assignment \citep{i2c}. We report our model performance when trained with only the mixture-EM optimization with both original and augmented images (Eq.~\eqref{eq:em_loss_aug}) and when trained with the two-fold optimization (Algorithm~\ref{alg:em}). For each dataset, we report the average accuracy of our algorithm over six trials. For the two-fold optimization, we also report the margin of error, considering 95\% confidence interval.

When trained with only the mixture-EM optimization for both original and transformed images, our models surpass all traditional clustering algorithms and existing deep clustering methods, which still rely on k-means such as Deep Embedded Clustering (DEC) \citep{dec} and DeepCluster \citep{deepcluster}. The two-fold optimization further improves the performance of the mixture-EM optimization, surpassing end-to-end deep clustering methods such as DECCA\citep{diallo2021deep}, SCAE \citep{scae}, DAC \citep{dac}, ADC \citep{adc} and IIC \citep{i2c} in most cases. IIC \citep{i2c} uses both labeled and unlabelled spaces of STL10 and, when trained with only the labeled portion, achieves only \textbf{49.9\%}. Furthermore, IIC uses many augmented samples in a batch by repeated augmentation (5 times). With augmentation once per batch, they only achieve \textbf{47\%} for the STL10 dataset. Also, IIC uses multiple heads with over-clustering strategies to improve overall performance.

\begin{figure}[ht]
	\begin{center}
	\includegraphics[width=0.8\linewidth]{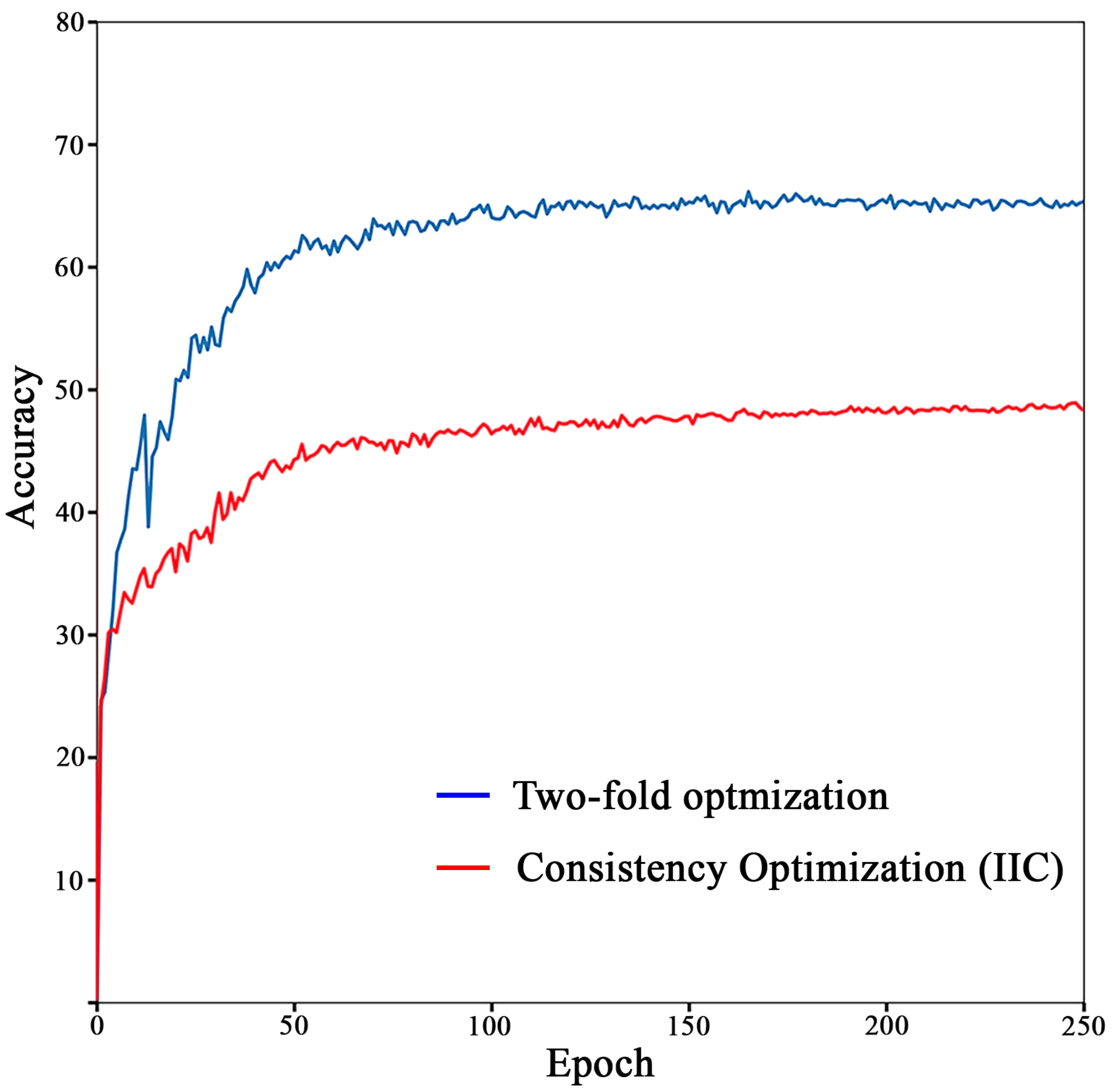}
	\end{center}
	\vspace{-0.2in}
	\caption{Training curves in STL10 dataset. The blue curve shows the two-fold optimization, and the red curve shows the consistency optimization in IIC \citep{i2c} in our setting. Our two-fold optimization, which utilizes mixture-EM optimization alongside consistency optimization, shows a significant improvement over consistency optimization alone.}
	\label{fig:stl_acc}
\end{figure}

In contrast, our framework reports an impressive \textbf{57.93\%} when trained with only the labeled portion, with lesser image augmentation, which is imposed only once per batch. We show that IIC with only one head achieves inferior performance in our training setting with lesser augmentation to both mixture-EM optimization and the two-fold optimization. The consistency optimization term in the two-fold optimization shares the same intuition to IIC, maintaining a similar model response to the original and its transformed images. Hence, we can conclude that our two-fold optimization improves sole consistency optimization in clustering. Figure \ref{fig:stl_acc} further shows the learning curves of IIC in our setting and the two-fold optimization, validating this fact.

It is important to note that certain deep image clustering methods report superior performance to our method \citep{van2020scan, wu2019dccm, park2021ruc, han2020mitigating}. However, most approaches are multi-stage methods consisting of initialization methods, multiple losses, and fine-tuning methods. For example, most of the performance improvement of SCAN can be attributed to the pre-text task learned prior to the clustering. In addition, SCAN also uses self-labeled fine-tuning. While our mixture-EM formulation optimization can also be extended with such pre-text learning, heavier augmentation, fine-tuning such as self-labeling, we omit such additions in this paper.

\subsection{Visualizations}
\label{ss:visualizations}

\begin{figure*}[ht]
	\begin{center}
	\begin{subfigure}{0.32\linewidth}
		\includegraphics[width=\linewidth]{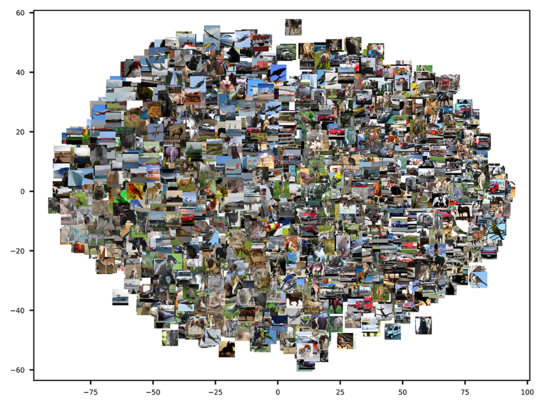}
		\caption{Random Network}
		\label{fig:tsne_random}
	\end{subfigure}
	\begin{subfigure}{0.32\linewidth}
	    \includegraphics[width=\linewidth]{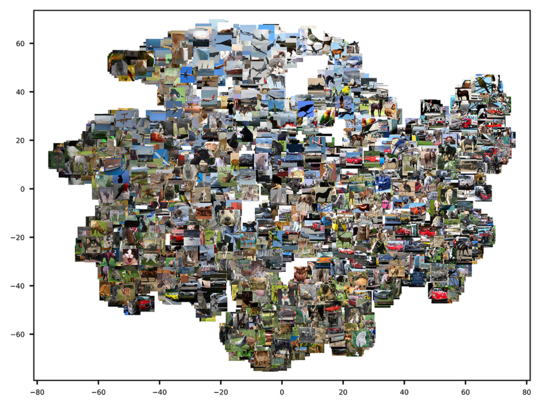}
		\caption{No Normalization}
		\label{fig:tsne_nobn}
	\end{subfigure}
	\begin{subfigure}{0.32\linewidth}
		\includegraphics[width=\linewidth]{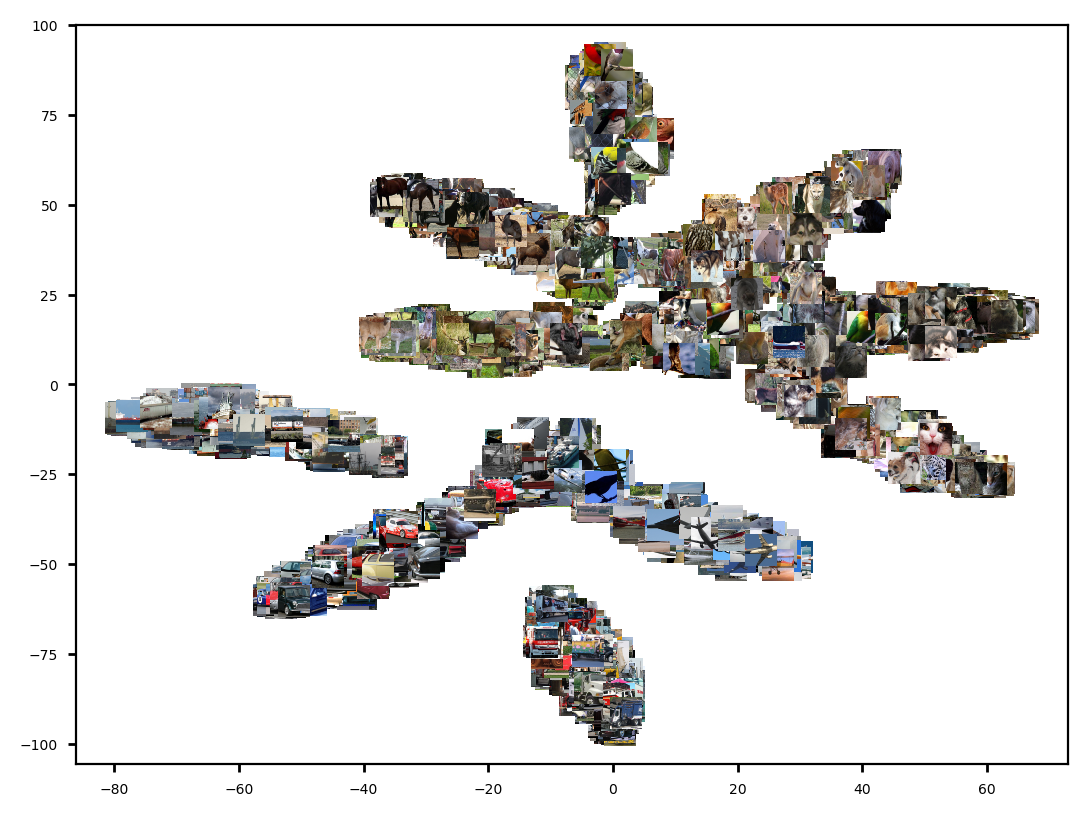}
		\caption{Normalization}
		\label{fig:tsne_bn}
	\end{subfigure}
	\end{center}
	\vspace{-0.2in}
	\caption{2-D mapping of the final-layer response for a subset of STL10. a) Random network: shows no knowledge of possible clusters. b) When trained without normalizing cluster relevance scores: collapses to trivial solutions (here, only one cluster). c) When trained with normalizing relevance scores: non-trivial observable meaningful clusters.}
	\label{fig:tsne}
	\vspace{-0.1in}
\end{figure*}

This section analyzes an STL10-trained network to validate the rich cluster modeling and feature extraction empirically. We first plot the network response before $\mathrm{softmax}$ in 2-d for a subset of STL10 containing 2560 images in Fig.~\ref{fig:tsne}. We use the T-SNE algorithm \citep{maaten_tsne} to map the network response vectors to 2-d while preserving the relationship between vectors. Fig.~\ref{fig:tsne_random} shows such visualization for a randomly initialized network. Fig.~\ref{fig:tsne_nobn} shows the network response when trained with the two-fold optimization, but without normalizing the cluster relevance scores over the batch. Fig.~\ref{fig:tsne_bn} shows the response when trained with the same loss and batch-normalization of cluster relevance scores.  The randomly initialized network (Fig.~\ref{fig:tsne_random}) contains no information on a possible clustering basis. Our algorithm trains such a network to categorize the sample space into meaningful clusters with observable cluster boundaries (Fig.~\ref{fig:tsne_bn}). Fig.~\ref{fig:tsne_nobn} shows the trivial convergence where a single cluster is formed with other clusters having no members. Fig.~\ref{fig:tsne_nobn} and Fig.~\ref{fig:tsne_bn} further show that normalizing relevance scores over the batch leads to a boost of performance and prevents the network from converging to trivial solutions, without any other refining technique such as normalizing by cluster assigned frequencies \citep{dec}.


\begin{figure*}[!ht]
    \captionsetup[subfigure]{labelformat=empty}
	\begin{center}
	\begin{subfigure}{0.83\linewidth}
		\includegraphics[width=\linewidth]{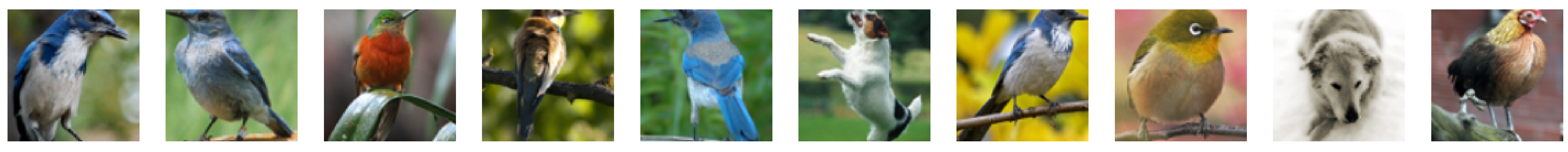}
		\caption{cluster 3}
	\end{subfigure}
	\begin{subfigure}{0.15\linewidth}
			\includegraphics[width=\linewidth]{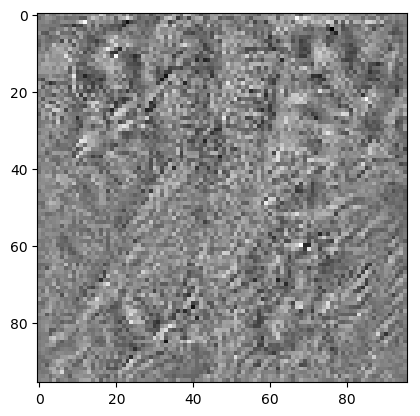}
	\end{subfigure}
	\begin{subfigure}{0.83\linewidth}
		\includegraphics[width=\linewidth]{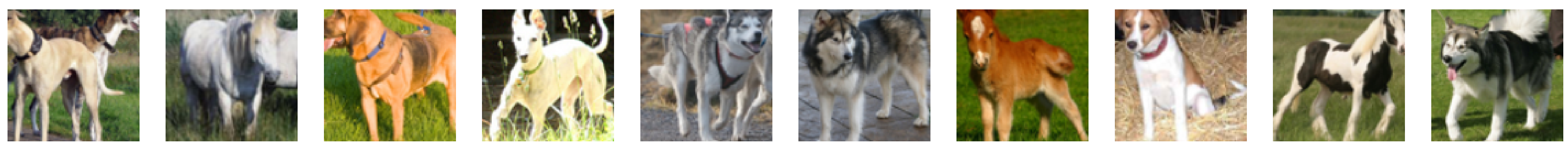}
		\caption{cluster 4}
	\end{subfigure}
	\begin{subfigure}{0.15\linewidth}
			\includegraphics[width=\linewidth]{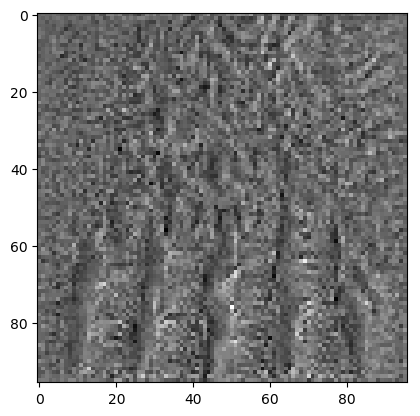}
	\end{subfigure}
	\begin{subfigure}{0.83\linewidth}
		\includegraphics[width=\linewidth]{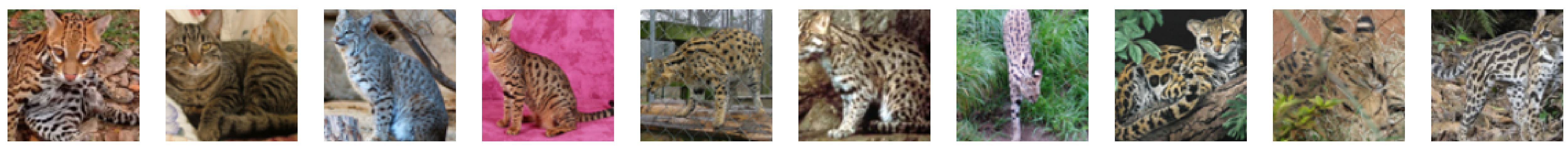}
		\caption{cluster 7}
	\end{subfigure}
	\begin{subfigure}{0.15\linewidth}
			\includegraphics[width=\linewidth]{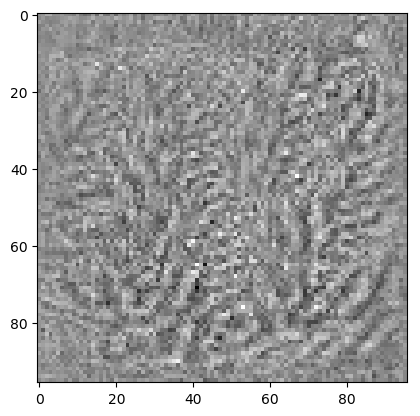}
	\end{subfigure}
	\begin{subfigure}{0.83\linewidth}
		\includegraphics[width=\linewidth]{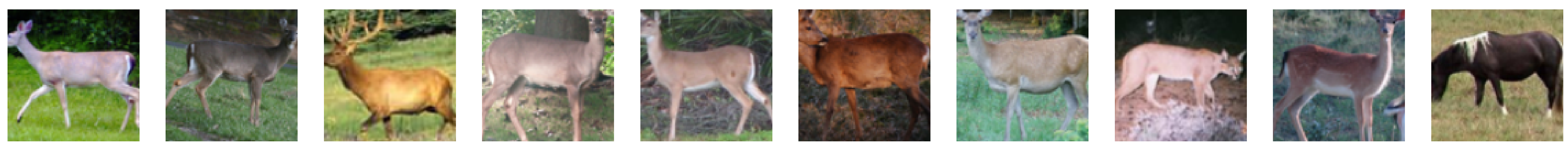}
		\caption{cluster 8}
	\end{subfigure}
	\begin{subfigure}{0.15\linewidth}
			\includegraphics[width=\linewidth]{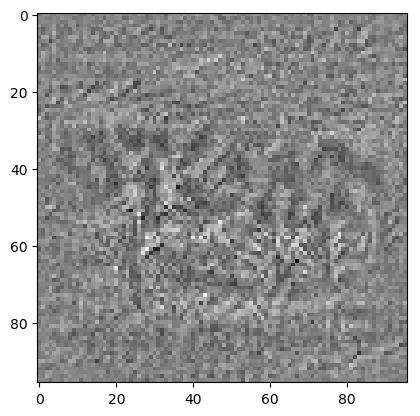}
	\end{subfigure}
	\begin{subfigure}{0.83\linewidth}
		\includegraphics[width=\linewidth]{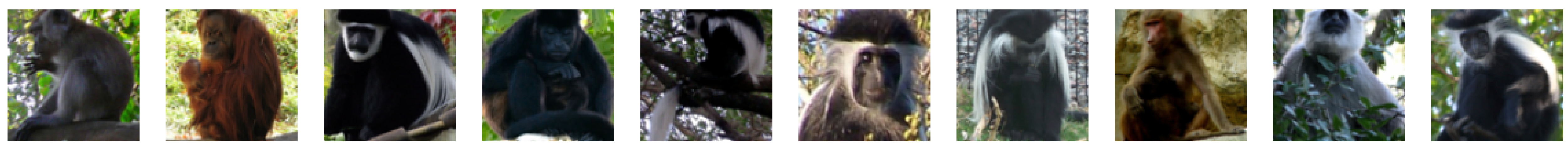}
		\caption{cluster 9}
	\end{subfigure}
	\begin{subfigure}{0.15\linewidth}
			\includegraphics[width=\linewidth]{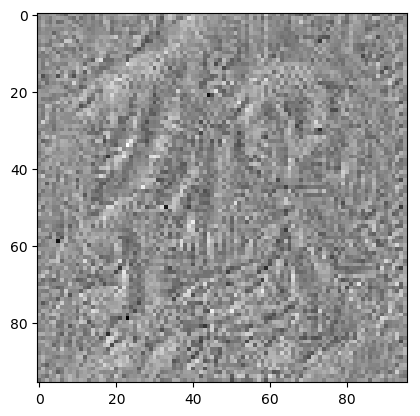}
	\end{subfigure}
	\end{center}
	\vspace{-0.2in}
	\caption{Five chosen clusters represented by the corresponding five neurons in the final layer for STL10 trained NN. Each row shows the ten images with the highest relevance score for the corresponding cluster, followed by the synthesized image that maximizes the particular score. Images with similar high-level information have been categorized together. The synthesized images show matching body structures and patterns.}
	\label{fig:best10}
	\vspace{-0.1in}
\end{figure*}

In Fig.~\ref{fig:best10}, we plot the ten images which output the highest values of the corresponding relevance scores for chosen five clusters, along with the synthesized image that maximizes the corresponding neuron. We construct the synthesized input by performing gradient ascent on the randomly initiated input image to maximize the corresponding node's response before the activation \citep{act_max}. The resulting visualizations illustrate that the model clusters images with similar abstract information together. The synthesized images match the corresponding high-level information contained in top member images. For example, cluster 4 mostly activates for dogs, and in the synthesized image, we can observe matching leg patterns. Cluster 7's best images mostly contain cats, and the synthesized image shows dotted patterns present in all ten highest activated images. Cluster 8's top images mostly show deers observed from the side, and the synthesized image shows matching body structure. This experiment validates the trained network's ability to model rich clusters end-to-end than limited hand-designed cluster characteristics.

\begin{figure}[!ht]
	\begin{center}
	\begin{subfigure}{\columnwidth}
	    \includegraphics[width=0.24\linewidth]{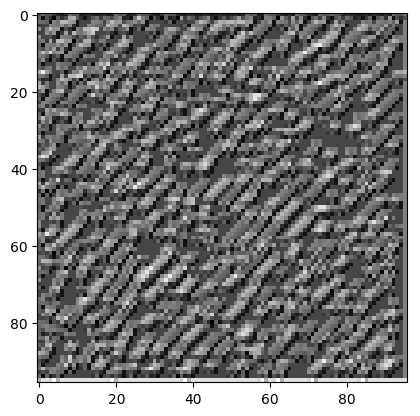}
	    \includegraphics[width=0.24\linewidth]{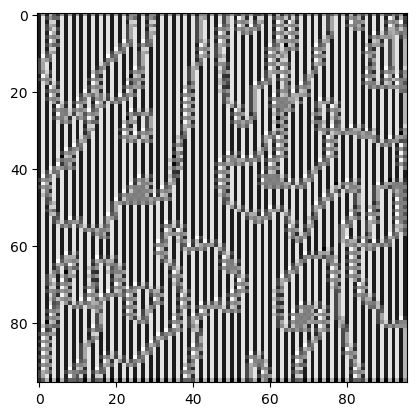}
		\includegraphics[width=0.24\linewidth]{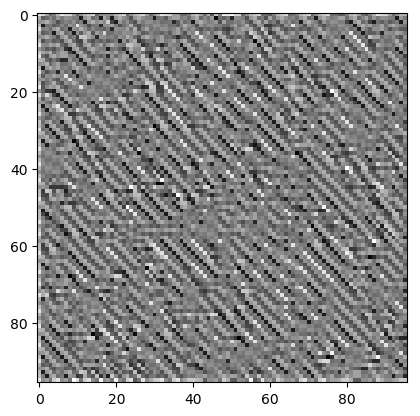}
		\includegraphics[width=0.24\linewidth]{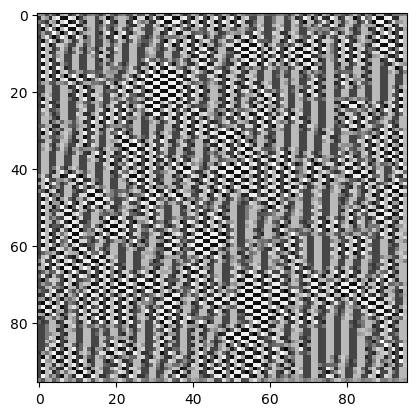}
		\caption{Initial convolutional layer filters}
		\label{sfig:filters_shallow}
	\end{subfigure}
	\begin{subfigure}{\columnwidth}
	    \includegraphics[width=0.24\linewidth]{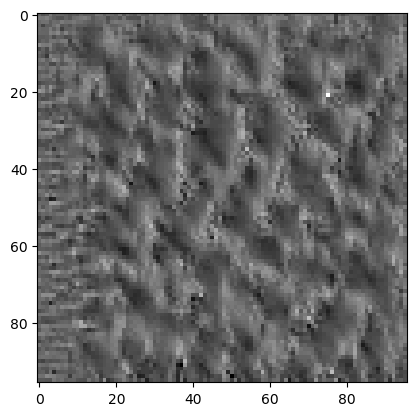}
	    \includegraphics[width=0.24\linewidth]{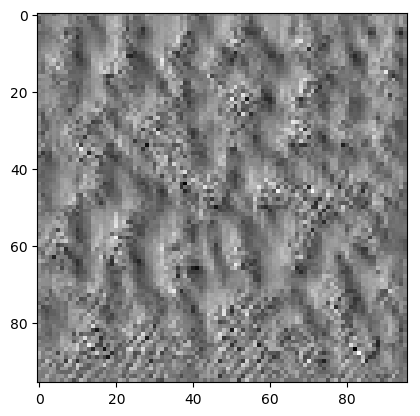}
		\includegraphics[width=0.24\linewidth]{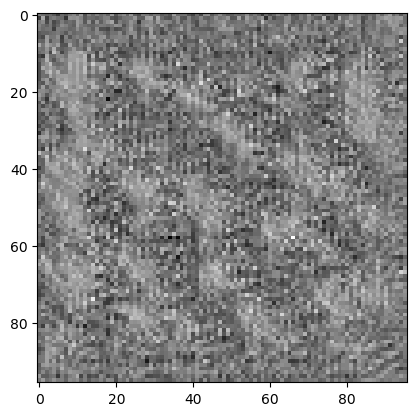}
		\includegraphics[width=0.24\linewidth]{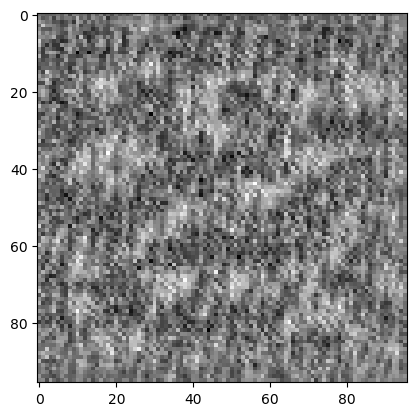}
		\caption{Final convolutional layer filters}
		\label{sfig:filters_deep}
	\end{subfigure}
	\end{center}
	\vspace{-0.2in}
	\caption{Filter visualization for the initial and the last convolutional layers. The initial layer shows simple features whereas, the deep filters show complex patterns. Overall, the convolutional filters are learning features, ascending in the level of abstraction with our unsupervised algorithm.}
	\label{fig:filters}
	\vspace{-0.1in}
\end{figure}

Finally, we visualize the convolutional filters of the model to observe and validate the convolutional feature extraction. To visualize a filter, we optimize a randomly initialized input to maximize the output of the particular filter \citep{act_max}. Fig.~\ref{fig:filters} gives such visualizations for initial layer convolutions and final layer convolutions. In each layer, we plot the synthesized images for four chosen filters. The shallow initial filters learn to extract low-level patterns (Fig.~\ref{sfig:filters_shallow}) and the deepest convolutional filters learn high-level patterns (Fig.~\ref{sfig:filters_deep}), i.e., the CNN learns features that are distributed along with the network depth with increasing complexity. This visualization proves that our fully unsupervised learning algorithm enables the convolutional filters to extract relevant patterns to the clustering task. Maintaining the model response to transformed images through consistency optimization enables this general feature extraction of convolutions.

\section{Conclusion}
\label{se:conclusion}
Our batch-wise mixture-EM formulation trains a neural network end-to-end to concurrently learn the cluster distributions and cluster assignment in an online fashion. It is a better alternative to k-means; we can replace k-means with a neural network and EM optimization in any scenario in the clustering context. 1) To efficiently cluster a set of raw data points to a given number of categories. 2) In image clustering context, as the clustering method to be used alongside representation learning or other techniques of general feature learning. The normalization of the cluster relevance scores over batches enables the sigmoid of these relevance scores to approximate the cluster distributions as PDFs of the observation, thus preventing trivial solutions. The visualizations empirically validated the meaningful cluster modeling, the rich convolutional feature extraction, and the effect of the relevance score normalization. Using a neural network of the required depth and the simplicity of the training process makes our algorithm easy to use in any form of clustering task with varying cluster modeling complexity. This paper presents results without support from heavy data augmentation, other representation learning techniques such as pre-text tasks, deeper networks or initialization, or fine-tuning methods. Nevertheless, it is worthwhile to investigate the possibility of further improvements by studying them. While we propose the $\mathrm{sigmoid}$ activation along with normalized relevance scores to model cluster distributions, it would be interesting to explore more sophisticated activations or methods to derive better distributions.

\section{Acknowledgement}
This work is funded by CODEGEN International (Pvt) Ltd, Sri Lanka.

\balance
\bibliographystyle{elsarticle-num} 
\bibliography{mybib}

\end{document}

%% file: sigmoids.tex
        \pgfplotsset{every tick label/.append style={font=\scriptsize}}
        \begin{tikzpicture}[>=latex]
        \begin{axis}[
             x = 0.32cm,
             y = 2.5cm,
            domain=-12:12,
            no markers,
            axis x line=center,
            axis y line=center,
            xlabel style={right},
            ylabel style={above},
            xlabel = {\scriptsize $a^\ast$},
		xtick={-11.225, -5, 0, 5, 11.225},
            xmin=-12.4,
            xmax=12.4,
            ymin=0,
            ymax=1.1,
            samples=100,
		legend style={
		at={(0.05,1)},
		draw=none,
		anchor=north west}            
            ]
            \addplot[thick, color=blue!70!black] {1/(1+exp(-x))};
 		\addplot[thick, color=red!70!black] {1/(1+exp(-x/5))};   
 		 \addplot[thick, dashed, color=orange!70!black] {1/(2.506)*exp(-1/2*(x^2)))};     
			\legend{\scriptsize $\mathrm{sigmoid}(a^\ast)$, \scriptsize $\mathrm{sigmoid}(a^\ast/5)$, \scriptsize  $\frac{1}{\sqrt{2\pi}}e^{-\frac{1}{2}x^2}$} 		 
        	\draw[green!70!black, dashed, thick] (axis cs:11.225,0) -- (axis cs:11.225,1.1);
        	\draw[green!70!black, dashed, thick] (axis cs:-11.225,0) -- (axis cs:-11.225,1.1);        	
        \end{axis}
    \end{tikzpicture}